%% file: main.tex
\def\year{2022}\relax
\documentclass[letterpaper]{article} 
\usepackage{aaai22}  
\usepackage{times}  
\usepackage{helvet}  
\usepackage{courier}  
\usepackage[hyphens]{url}  
\usepackage{graphicx} 
\usepackage{natbib}  
\usepackage{caption} 
\DeclareCaptionStyle{ruled}{labelfont=normalfont,labelsep=colon,strut=off} 
\usepackage{algorithm}
\usepackage{algorithmic}

\usepackage{amsmath} 
\usepackage{amssymb}
\usepackage{amsthm}

\usepackage{xcolor}
\usepackage{subcaption}
\usepackage{breqn}
\usepackage{subfiles}
\usepackage{microtype}
\usepackage{multirow}
\usepackage{booktabs}
\usepackage{bbm}
\usepackage[group-separator={,}]{siunitx}

\usepackage{newfloat}
\usepackage{listings}

\newtheorem{theorem}{Theorem}
\newtheorem*{theorem*}{Theorem}

\newtheorem{lemma}{Lemma}
\newtheorem*{lemma*}{Lemma}
\newtheorem{assumption}{Assumption}

\newcommand{\argmin}{\operatornamewithlimits{arg\,min}}
\newcommand*{\blue}{\textcolor{blue}}

\newcommand{\sysname}{\textsc{NestedMAML}}
\newcommand{\biopt}{nested bi-level}

\DeclareMathAlphabet{\mathcal}{OMS}{cmsy}{m}{n} 

\floatstyle{ruled}
\newfloat{listing}{tb}{lst}{}
\floatname{listing}{Listing}
\title{A Nested Bi-level Optimization Framework for Robust Few Shot Learning}

\author{
    Krishnateja Killamsetty\equalcontrib,
    Changbin Li\equalcontrib,
    Chen Zhao,
    Rishabh Iyer,
    Feng Chen
}
\affiliations{

    Department of Computer Science\\
    The University of Texas at Dallas \\
    Richardson, Texas, USA\\
    \{krishnateja.killamsetty, changbin.li,  chen.zhao, rishabh.iyer, feng.chen\}@utdallas.edu
}

\begin{document}

\maketitle

\begin{abstract}
\input{abstract}
\end{abstract}

\section{Introduction}
\input{introduction_v2}

\section{Preliminaries}
\input{problem_formulation}

\section{Methodology}
\input{algorithm}

\section{Experiments}
\input{experiments}

\section{Conclusion}
\input{conclusion}

\bibliography{main}

\bigskip
\appendix
\input{appendix}
\end{document}

%% file: abstract.tex
Model-Agnostic Meta-Learning (MAML), a popular gradient-based meta-learning framework, assumes that the contribution of each task or instance to the meta-learner is equal. Hence, it fails to address the domain shift between base and novel classes in few-shot learning. In this work, we propose a novel robust meta-learning algorithm,  \sysname{}, which learns to assign weights to training tasks or instances. We consider weights as hyper-parameters and iteratively optimize them using a small set of validation tasks set in a \biopt{} optimization approach (in contrast to the standard bi-level optimization in MAML). We then apply \sysname{} in the meta-training stage, which involves (1) several tasks sampled from a distribution different from the meta-test task distribution, or (2) some data samples with noisy labels. Extensive experiments on synthetic and real-world datasets demonstrate that \sysname{} efficiently mitigates the effects of "unwanted" tasks or instances, leading to significant improvement over the state-of-the-art robust meta-learning methods.

%% file: introduction_v2.tex
Meta-learning~\citep{schmidhuber1987, Naik1992MetaneuralNT,santoro2016meta, vinyals2016matching, finn2017model} can achieve quick adaption for UNSEEN tasks by identifying common structures among various SEEN tasks, enabling faster learning of a new task with as little data as possible. 
However, existing meta-learning techniques (\emph{e.g.,} MAML~\citep{finn2017model}) often fail to generalize well when the test tasks belong to a different distribution from the training tasks distribution~\citep{Chen2019ICLR}. 
For example, MAML assumes equal weights to all samples and tasks during meta-training.
This task homogeneity assumption of MAML often limits its ability to work in real-world applications. 
\begin{figure}[!t]
    \centering
    \includegraphics[width=0.47\textwidth, height=4.2cm]{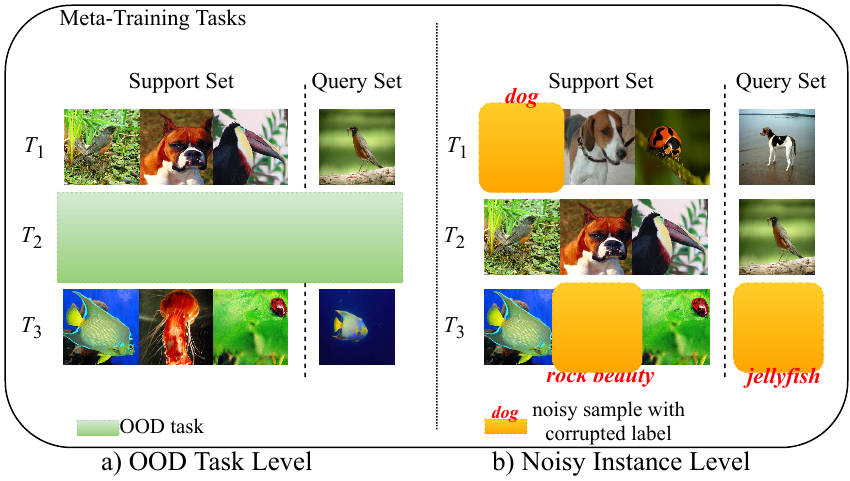}
    \vspace{-7mm}
    \caption{We consider corrupted training set for few-shot learning in this work: a) OOD task level and b) noisy instance level. $T_2$ in a) is an OOD task that is sampled from a different distribution. b) contains some noisy samples which are mislabeled. For example, the actual label of the first sample in $T_1$ should be ``\textit{Arctic fox}" which is labeled as ``\textit{dog}"; The labels of two noisy samples in $T_3$ are flipped wrongly. The first one should be ``\textit{rock beauty}," and the other one should be ``\textit{jellyfish}."}
    \label{fig:motivation}
\vspace{-8mm}
\end{figure}

\begin{figure*}[!htbp]
    \centering
    \includegraphics[width=0.85\linewidth, height=6.5cm]{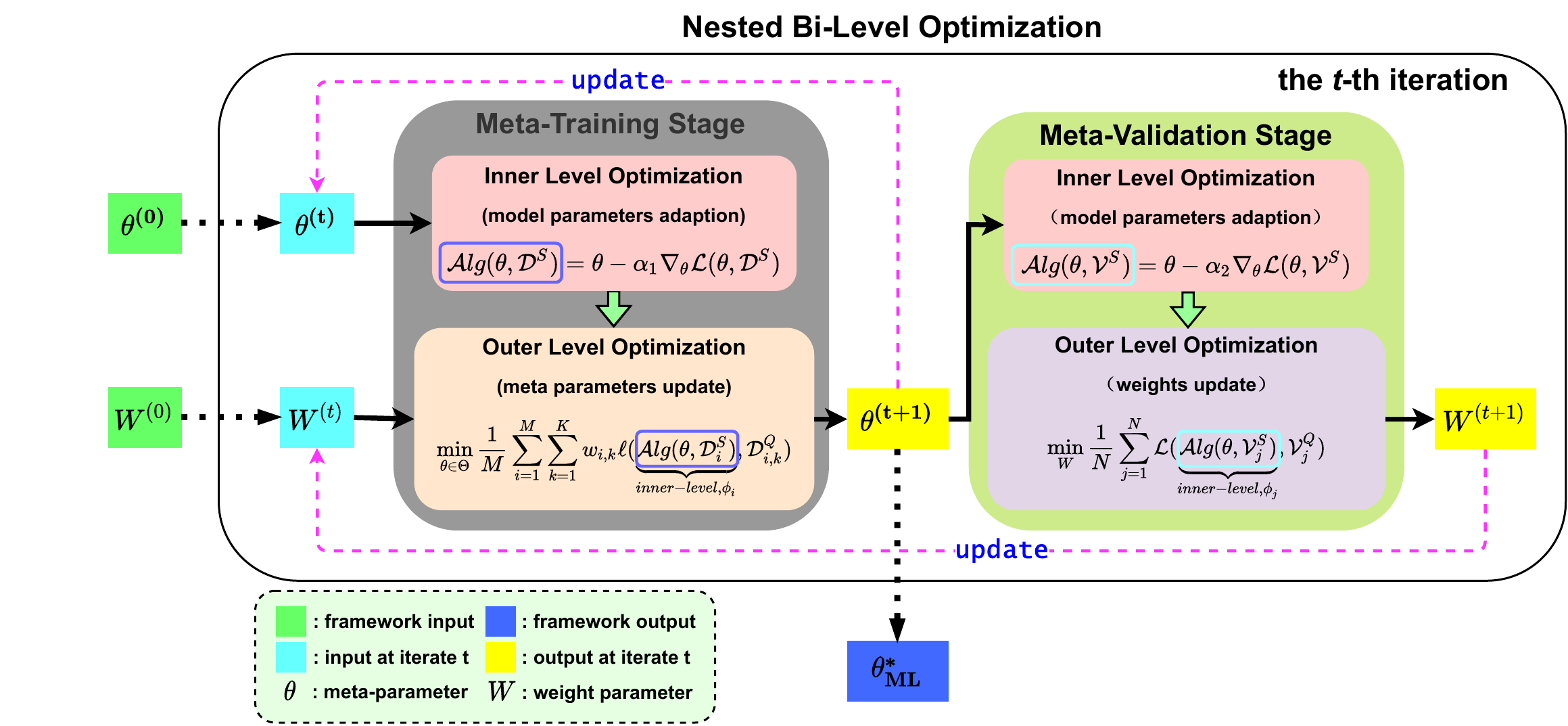}
    \caption{Overview of our \sysname{} framework that solves a \textit{\biopt{}} optimization problem. (a) In the meta-training stage, model parameters $\phi_i$ of each task are adapted from meta-parameter $\boldsymbol{\theta}$ through the inner level optimization; (b) In the outer-level of the meta-training stage, we update the meta-parameters using the weights $W$ from the previous iterate; (c) Weights are further updated in the meta-validation stage using the gradient of the meta-losses with respect to current $W$.} 
    \label{fig:overview}
    \vspace{-4mm}
\end{figure*}

We motivate the importance of robust meta-learning when meta-training tasks have OOD tasks using the following examples. For example, consider the task of detecting vehicles at night under different weather conditions. In this case, the meta-test tasks only consist of images of vehicles at night. Since the procurement vehicles driving data at night, covering all critical scenarios is difficult, we need a model that can quickly adapt to rare driving conditions. Hence, we consider meta-training tasks to consist of images of the vehicles in multiple lighting scenarios. In this case, some of the tasks in meta-training may degrade the meta-test performance. So, it is vital to have a meta-learning model that is robust to OODs.

Another example is rare lung cancer detection from medical x-ray images. Since the procurement of rare lung cancer images is both problematic and expensive, it is beneficial to use prior knowledge of cancer images. Specifically, the meta-test tasks contain images of rare lung cancer, whereas meta-training tasks consist of general cancer x-ray images. In the examples given above, meta-test tasks belong to specialized slices where the data availability is meager compared to meta-training tasks. The meta-training task distribution is biased compared to that in the meta-test. To keep the whole meta-training tasks for generalization and reduce the adverse impact of the biased distribution in meta-training, we propose a novel robust few-shot learning algorithm in the presence of outliers in meta-training time, which is similar to the corruptions in training time in the traditional robust learning~\citep{schneider2020improving}. This is different from the existing robust few-shot learning papers~\cite{yin2018adversarial, lu2020robust, goldblum2020adversarially} which consider the corruption only happens in meta-test time.     

To simulate the corruptions in meta-training, two levels of outliers (Figure \ref{fig:motivation}) are considered: {\textit{a}}) Out-Of-Distribution (OOD) task, where the meta-training has tasks that are out of distribution to the meta-test tasks (i.e., the meta-test dataset is a specialized slice of meta-train) and {\textit{b}}) noisy instance level, where some of the labels might be noisy (due to human labeling errors or inherent ambiguity of certain classification problems) for meta-training samples. 

A natural way of dealing with corrupted data in meta-training is by assigning weights to either tasks or individual instances. For example, assigning zero weight to OOD or noisy tasks/instances in the meta-train set improves the meta-learning algorithm's performance. Inspired by~\citep{ren2018learning}, in this work, we propose an end-to-end robust meta-learning framework called \sysname{} that can achieve the reweighting schema along with learning good model initialization parameters in the few-shot learning scenario.  




\sysname{} considers the weights as hyper-parameters and uses a small set of meta-validation tasks representing the meta-test tasks to find the optimal hyper-parameters by minimizing the meta-loss on the validation tasks in a \textbf{\biopt{}} manner.  An overview of \sysname{} is given in Figure \ref{fig:overview}. In practice, the size of the meta-validation tasks set required by \sysname{} is tiny compared to the meta-training dataset. Hence, creating a  small and clean meta-validation set is neither expensive nor unrealistic, even for rare specialized use cases of a real-life scenario. A similar strategy has been applied in ~\citep{ren2018learning,shu2019meta,killamsetty2020glister}. However, they focus on traditional supervised learning, and we generalize this to task- and instance-level in a meta-learning setting. Since \sysname{} uses an online framework to perform a joint optimization of the weight hyper-parameters and model parameters for the weighted MAML model, the computational time of ours is comparable to MAML.

\noindent \textbf{Contributions of our work are summarized as follows:} 1) We study the general form of the task and instance weighted meta-learning, where we learn the optimal weights and model initialization parameters by optimizing a \textit{\biopt{}} objective function. To the best of our knowledge, ours is the first work that studies the \textit{\biopt{}} optimization problem, which comes naturally in such a new setting. 2) We introduce a novel algorithmic framework \sysname{} that uses a small set of validation tasks to enable robust meta-learning. We solve the \emph{\biopt{}} optimization problem efficiently through a series of practical approximations and provide a theoretical convergence analysis for \sysname{}. In particular, we show that \sysname{} converges in $\mathcal{O}(1/\epsilon^2)$ iterations under reasonable assumptions and contrast this with existing bounds of MAML. 3) We provide comprehensive synthetic and real-world data experiments demonstrating that \sysname{} achieves state-of-the-art results in two scenarios (OOD tasks and noisy instance labels).

\vspace{-1ex}
\section{Related Work}
There are several lines of meta-learning algorithms: nearest neighbors-based methods~\citep{vinyals2016matching}, recurrent network-based methods~\citep{ravi2016optimization}, and gradient-based methods. As the representative of gradient-based meta-learning algorithms, MAML~\citep{finn2017model} and its variants ~\citep{finn2018probabilistic, nichol2018first, rusu2018meta,rajeswaran2019meta, behl2019alpha, raghu2019rapid, zhao2020fair, zhou2020meta} learn a shared initialization of model parameters across a variety of tasks during the meta-training phase that can adapt to new tasks using a few gradient steps.~\citet{cai2020weightedmeta} proposes a simple weighted meta-learning approach only for the basis regression problem that selects weights by minimizing a data-dependent bound involving an empirical integral probability metric between the weighted sources and target risks. However, this approach cannot be easily extended to complex scenarios with arbitrary loss functions.

There are few meta-learning papers discussing learning with OOD tasks. ~\citet{jeong2020ood} propose an OOD detection framework in meta-learning through generating fake samples which resemble in-distribution samples and combine them with real samples. However, they assume the outlier instances exist in the query set, which is different from ours. The most relevant field is from the perspective of task heterogeneity~\citep{vuorio2019multimodal, triantafillou2019meta,yao2020automated}. ~\citet{vuorio2019multimodal} proposed  MMAML to deal with multimodal task distribution with disjoint and far apart modes and generates a set of separate meta-learned prior parameters to deal with each mode of a multimodal distribution. If we view that all the OOD tasks belong to a single mode, this is relevant to our setting. To tackle the distribution drift from meta-training to meta-test, B-TAML~\citep{lee2020l2b} learn to relocate the initial parameters to a new start point based on the arriving unseen tasks in the meta-test. The setting considered in our work and B-TAML work can be viewed as similar if we assume some of the datasets considered in the multi-dataset classification setting of B-TAML as OOD datasets.

To tackle samples with corrupted labels, some researches~\citep{luo2015foveation, jalal2017robust, wang2019direct} introduce noise-robust models.~\citet{ren2018learning} and~\citet{shu2019meta} propose a noisy data filtering strategy using an instance reweighting strategy where the weights are learned automatically. However, the effect of noisy labels on few-shot learning requires more attention. Although ~\citep{yin2018adversarial,lu2020robust,goldblum2020adversarially} proposes robust meta-learning or few-shot learning, they assume a presence of outliers containing in meta-test, which is different from ours. 



%% file: problem_formulation.tex
\subsection{Notations}
In the setting of meta-learning for few-shot learning, there is a set of meta-training tasks $\{\mathcal{T}_i\}_{i=1}^M$ sampled from the probability distribution 
$p_{tr}(\mathcal{T})$. Each few-shot learning task $\mathcal{T}_i$ has an associated dataset $\mathcal{D}_i$ containing two disjoint sets $\{\mathcal{D}^{S}_{i},\mathcal{D}^{Q}_{i}\}$, where the superscripts $S$ and $Q$ denote support set and query set respectively. The query sets take the form $\mathcal{D}^{Q}_{i} = \{\boldsymbol{x}_i^k, {y}_i^k\}_{k=1}^K$ and similarly for $\mathcal{D}^{S}_{i}$. Meta-validation tasks are denoted in a similar manner: $\{\mathcal{T}^{\mathcal{V}}_{j}= \{\mathcal{V}_{j}^{S}, \mathcal{V}_{j}^{Q}\} \}_{j = 1} ^ {N}$
Let the loss function be denoted as $\mathcal{L}(\boldsymbol{\phi}, \mathcal{D})$ with $\boldsymbol{\phi}$  denoting model parameters and $\mathcal{D}$ denoting the dataset, and $\ell(\boldsymbol{\theta}, d)$ with model parameters $\boldsymbol{\theta}$ on the data-point $d$. For example, $\mathcal{L}(\phi, \mathcal{D}_i^{Q})$ denotes the loss of the $i^{th}$ training task query set $\mathcal{D}_i^{Q}$ for given model parameters $\phi \in \boldsymbol{\Phi}\equiv \mathbb{R}^d$, where $\boldsymbol{\phi}:=\mathcal{A}lg(\boldsymbol{\theta}, \mathcal{D}^S)$ and $\boldsymbol{\theta}\in \boldsymbol{\Theta} \equiv \mathbb{R}^d$ is the meta-parameter. $\mathcal{A}lg(\cdot)$ corresponds to a learning algorithm.


For notation convenience, we write $\mathcal{L}_i(\phi) := \mathcal{L}(\phi, \mathcal{D}_i^{Q})$; $\mathcal{L}_{V_j}(\phi):=\mathcal{L}(\phi, \mathcal{V}_j^{Q})$;  $\widehat{\mathcal{L}}_{V_j}(\boldsymbol{\phi}) := \mathcal{L}(\boldsymbol{\phi}, \mathcal{V}_j^{S})$. We denote scalars by lower case italic letters, vectors by lower case boldface letters, and matrices by capital italic letters throughout the paper. A table of notations with corresponding explanations is given in Appendix~\ref{app:notations}.



\subsection{Model-Agnostic Meta-Learning}
The goal of MAML~\cite{finn2017model} is to obtain the optimal initial parameters that minimize the meta-training objective:
\begin{equation}
\begin{aligned}
    \label{meta-objective}
    \overbrace{\boldsymbol{\theta}^*_{ML}= \argmin_{\boldsymbol{\theta} \in \boldsymbol{\Theta}}{\mathcal{F}(\boldsymbol{\theta})}}^{outer-level}\text{\hspace{1.7cm}}\\
    \text{where, }\mathcal{F}(\boldsymbol{\theta}) = \frac{1}{M}\sum\nolimits_{i=1}^{M}\mathcal{L}(\overbrace{\mathcal{A}lg(\boldsymbol{\theta}, \mathcal{D}_{i}^{S})}^{inner-level}, \mathcal{D}_{i}^{Q})\text{\hspace{0.5cm}}
\end{aligned}
\end{equation}
This is a bi-level optimization problem, where we construe that $\mathcal{A}lg(\boldsymbol{\theta}, \mathcal{D}_{i}^{S})$ explicitly or implicitly optimizes the inner-level task-specific adaptation. The outer-level corresponds to the meta-training objective of generalizing well (i.e. low test error) on the query set of each task after adaptation. 
 
Since $\mathcal{A}lg(\boldsymbol{\theta}, \mathcal{D}_{i}^{S})$ corresponds to single or multiple gradient descent steps. In case of a single gradient descent, $\mathcal{A}lg(\boldsymbol{\theta}, \mathcal{D}_{i}^{S})$ can be perceived as follwing:
\begin{equation}
\label{param-adaptation}  \mathcal{A}lg(\boldsymbol{\theta}, \mathcal{D}_{i}^{S}) = \boldsymbol{\theta} - \alpha \nabla_{\boldsymbol{\theta}}\mathcal{L}(\boldsymbol{\theta}, \mathcal{D}_{i}^{S})
\end{equation}
where $\alpha $ is a learning rate.
As shown above, the meta-training objective assumes equal weights to each task for generalization, which may not be ideal in the case of adversaries in the training tasks set.

%% file: algorithm.tex
\subsection{Problem Formulation}
This section discusses a more generalized meta-learning framework, where we weigh all the data instances in the query set of a task. One of the significant purposes for considering weighted meta-learning is to make it more robust to adversaries during training.

In meta-learning, the support and query datasets $\{\mathcal{D}_{i}^{S}, \mathcal{D}_{i}^{Q}\}$ for each task $\mathcal{T}_{i}$ are usually sampled from an underlying dataset $\mathcal{D}$. In \textit{instance-level weighting}, we associate each data instance $\{\mathcal{D}_{ik}^{Q} \mid k\in [K]\}$ in the query set of task $\mathcal{T}_{i}$ with a particular weight $w_{ik}$, where $K$ is the number of datapoints (instances) in the query set $\mathcal{D}_{i}^{Q}$. 
The problem can be formulated as follows:
\begin{equation}
\small
    \begin{aligned}
    \label{instance-weighting}
         \boldsymbol{\theta}^*_{ML}= \argmin_{\boldsymbol{\theta} \in \boldsymbol{\Theta}} {\mathcal{F}_w(\boldsymbol{\theta})}\text{\hspace{-3cm}}\\
         \text{where\hspace{2mm}}\mathcal{F}_w(\boldsymbol{\theta}) &= \frac{1}{M}\sum\nolimits_{i=1}^{M} \sum\nolimits_{k=1}^{K}w_{ik}\ell(\mathcal{A}lg(\boldsymbol{\theta}, \mathcal{D}_{i}^{S}), \mathcal{D}_{ik}^{Q}) \\
         &=\frac{1}{M}\sum\nolimits_{i=1}^{M} \mathbf{w}_{i}\mathcal{L}_{i}(\mathcal{A}lg(\boldsymbol{\theta}, \mathcal{D}_{i}^{S})) 
    \end{aligned}
\end{equation}
In the expression above, 
$$
\small
\mathcal{L}_{i}(\mathcal{A}lg(\boldsymbol{\theta},\mathcal{D}_i^{S})) = \begin{bmatrix} \ell(\mathcal{A}lg(\boldsymbol{\theta},\mathcal{D}_i^{S}), \mathcal{D}_{i1}^{Q}) \\ \dots \\ \ell(\mathcal{A}lg(\boldsymbol{\theta},\mathcal{D}_i^{S}), \mathcal{D}_{ik}^{Q}) \\ \dots \\ \ell(\mathcal{A}lg(\boldsymbol{\theta},\mathcal{D}_i^{S}), \mathcal{D}_{iK}^{Q}) \end{bmatrix}
$$
and $\mathbf{w}_{i} = [w_{i1}, \dots, w_{iK}]$ is the weight vector corresponding to the query set of task $\mathcal{T}_i$. The \textit{instance-level weighting} is useful in the scenarios where our underlying dataset $\mathcal{D}$ is prone to noisy labeled instances where an appropriate instance-level weighting can be used to distinguish the noisy samples with corrupted labels in the task. An ideal weight assignment is assigning large weight values to clean samples and small weight values to noisy samples in a task. 
 
Likewise, we discuss a special case of the instance weighting scheme called \textit{task-level weighting}, where we assign equal weights to every instance in the query set of a single task. \textit{Task-level weighting} is applied in scenarios where every instance in a task's query set is from an OOD task distribution or an In-Distribution (ID) task. In this case, the optimal weight assignment assigns small weight values to an OOD task and large weight values to an ID task.
\vspace{-1ex}
\subsection{\textsc{\biopt{}} Optimization}
Since we do not know the optimal weight assignment for real-world datasets, we need to learn the weights before training the \textit{instance-level weighting} model using the bi-level optimization problem defined in Eq.\eqref{instance-weighting}.

\sysname{} solves for optimal weight assignments by posing them as hyper-parameters using the optimization problem defined in Eq.\eqref{inwt-opt}. As seen in the optimization equation, \sysname{} uses a clean held-out meta-validation task set $\{\mathcal{T}^{\mathcal{V}}_{j}= \{\mathcal{V}_{j}^{S}, \mathcal{V}_{j}^{Q}\} \}_{j = 1} ^ {N}$ that is assumed to be relevant to test task distribution for generalization performance. In practice, the meta-validation task set's size is small compared to that of the meta-training tasks set $(N \ll M)$.  Hence, \sysname{} tries to select the weight hyper-parameters minimizing the model's meta-validation loss after taking a few gradient steps from the initial model parameters set using the instance-level weighting scheme.

The weight optimization objective for the instance-weighted MAML schema is as follows:
\begin{equation}
\small
    \begin{aligned}
    \label{inwt-opt}
     {W}^{*} &= \argmin_{\mathbf{w}} \frac{1}{N} \sum\nolimits_{j=1}^{N}\mathcal{L}(\mathcal{A}lg(\boldsymbol{\theta}^{*}_{W}, \mathcal{V}_{j}^{S}), \mathcal{V}_{j}^{Q}) 
     \\
    \text{where\hspace{2mm}} \boldsymbol{\theta}^{*}_{W} &= \argmin_{\boldsymbol{\theta} \in \boldsymbol{\Theta}} \frac{1}{M}\sum\nolimits_{i=1}^{M}\mathbf{w}_{i}^{*}\mathcal{L}(\mathcal{A}lg(\boldsymbol{\theta}, \mathcal{D}_i^{S}), \mathcal{D}_i^{Q})
    \end{aligned}
\end{equation}
and ${W}=[\mathbf{w}_1,\dots,\mathbf{w}_M]^{\intercal}$. Since the optimization problem for $\boldsymbol{\theta}_{W}^*$ is a standard bi-level optimization problem (\textit{i.e.} MAML), the complete optimization problem (Eq.\eqref{inwt-opt}) turns out to be a \textbf{\biopt{}} optimization problem. It involves solving a standard bi-level optimization problem for every weight configuration, and hence naively solving this \textbf{\biopt{}} optimization problem is intractable. Hence, we adopt an online and one-step meta-gradient based approach to solve the optimization problem more efficiently. 

\vspace{-1ex}
\subsection{The \sysname{} Algorithm}
To reduce the optimization problem's {(Eq.\eqref{inwt-opt})} computation complexity, we solve the optimization problem in an iterative manner where we optimize the model parameters and weight hyperparameter by taking a single gradient step. This process is repeated until we reach convergence. Hence, we approximate the solution to the model parameters optimization in Eq.\eqref{inwt-opt} first by adapting to each task using a single gradient step towards the inner task adaptation objective's descent direction and then taking a single gradient step towards the meta objective's descent direction. 

Assuming that at every iterate $t$ of training, a mini-batch of training tasks $\{{\mathcal{T}_i} \mid 1 \leq i \leq m \}$ is sampled, where $m$ is the mini-batch size and $m \ll M$, the optimal model parameters update of the above problem is as follows: 
\begin{equation}
\small
    \begin{aligned}
    \label{online-equation}
    \boldsymbol{\theta}^{(t)}_{W} = \boldsymbol{\theta}^{(t)} - \eta \frac{1}{m}\sum\nolimits_{i=1}^{m} \mathbf{w}_i^{(t)} \nabla_{\boldsymbol{\theta}} \mathcal{L}_i(\mathcal{A}lg(\boldsymbol{\theta},\mathcal{D}_i^{S}))|_{\boldsymbol{\theta}^{(t)}}
    \end{aligned}
\end{equation}
where $\eta$ is meta objective's step-size and $\alpha$ is the inner objective's step-size. 
After this, the optimal weight optimization problem will be as follows:
\begin{equation}
\small
    \begin{aligned}
        W^{*} = {\operatorname*{arg\,min\hspace{1mm}}_{W}} \frac{1}{N} \sum\nolimits_{j=1}^{N}\mathcal{L}_{V_j}(\mathcal{A}lg(\boldsymbol{\theta}_{W}^{(t)},\mathcal{V}_j^{S}))
    \end{aligned}
\end{equation}
Similarly, we optimize the weight hyperparameters by taking a single gradient step towards the meta-validation loss descent. We want to evaluate the impact of training a model on the weighted MAML objective against the meta-objective of sampled validation tasks $\{\mathcal{T}_{j}^{V} \mid 1 \leq j \leq n \}$ where, $n$ is the mini-batch size and $n \ll N$.
The weight update equation for the instance weighting scheme is as follows:
\begin{equation}
\small
    \begin{aligned}
        {W}^{(t+1)} = {W}^{(t)} - \frac{\gamma}{n} \sum\nolimits_{j=1}^{n}\nabla_{W}\mathcal{L}_{V_j}(\mathcal{A}lg(\boldsymbol{\theta}^{(t)}_{W},\mathcal{V}_j^{S}))
    \end{aligned}
    \label{online-equation-weight}
\end{equation}
where $\gamma$ is the weight update's step size.
The Lemma below provides the gradient of the meta-validation loss $\frac{1}{n} \sum_{j=1}^{n}\nabla_{W}\mathcal{L}_{V_j}(\mathcal{A}lg(\boldsymbol{\theta}^{(t)}_{W},\mathcal{V}_j^{S}))$ w.r.t. the weight vector $\mathbf{w}_{i}$, therefore giving the full update equation.
\begin{lemma}
\label{weight-update-lemma} 
The weight update for an individual weight vector $\mathbf{w}_{i}$ of the task $\mathcal{T}_i$ from time step $t$ to  $t+1$ is as follows:
\begin{align}
\small
\label{weight-update}
    \mathbf{w}_i^{(t+1)}&=\mathbf{w}_i^{(t)} + \frac{\eta \gamma}{mn} \sum\nolimits_{j=1}^{n} \nabla_{\phi_j} \mathcal{L}_{V_j}  \Big(\nabla_{\boldsymbol{\theta}}\mathcal{L}_i(\mathcal{A}lg(\boldsymbol{\theta}, \mathcal{D}_i^S))^{\intercal} \nonumber\\
    &- \alpha  \nabla^2\widehat{\mathcal{L}}_{V_j}|_{\boldsymbol{\theta}_{W}^{(t)}} \nabla_{\boldsymbol{\theta}}\mathcal{L}_i(\mathcal{A}lg(\boldsymbol{\theta}, \mathcal{D}_i^S))^{\intercal} \Big) 
\end{align}
where $\phi_j=\mathcal{A}lg(\boldsymbol{\theta}, \mathcal{V}_{j}^{S})$.
\end{lemma}
The proof is in Appendix \ref{app:Lemma1proofs}.
Once the optimal weights $\mathbf{w}^{(t+1)}$ at $t+1$ are achieved, we train the model using the new weights:
\begin{equation}
\small
    \begin{aligned}
        \boldsymbol{\theta}^{(t+1)} = \boldsymbol{\theta}^{(t)} - \frac{\eta}{m}\sum\nolimits_{i=1}^{m} \mathbf{w}_{i}^{(t+1)}\nabla_{\boldsymbol{\theta}}\mathcal{L}_i(\mathcal{A}lg(\boldsymbol{\theta}^{(t)}, \mathcal{D}_i^{S}))
\end{aligned}
    \label{model-update}
\end{equation}
We repeat the steps given in the equation \eqref{online-equation} from $t=1$ until convergence. See Algorithm~\ref{alg-RWMAML} for the full pseudo-code of \sysname.

\noindent \textbf{First-Order Approximation (\sysname{}-FO).} Even after the one step gradient approximation, the weight gradient calculation involves calculating multiple Hessian vector products, which is expensive. Since the coefficient of the Hessian vector-product term in the weight update (Eq.~\eqref{weight-update}) involves the product of three learning rate terms $\eta\alpha\gamma$,  we can make an approximation that the term involving the Hessian vector-product term is close to 0, given that the above learning rates are small. 
The approximated weight update takes the following form:
\begin{equation}
\small
    \label{weight-update-appr}
        \mathbf{w}_i^{(t+1)}=\mathbf{w}_i^{(t)} + \frac{\eta \gamma}{mn} \sum\nolimits_{j=1}^{n} \nabla_{\phi_j} \mathcal{L}_{V_j} \nabla_{\boldsymbol{\theta}}\mathcal{L}_i(\mathcal{A}lg(\boldsymbol{\theta}, \mathcal{D}_i^S))^{\intercal}
\end{equation}
This approximation is similar to the first-order approximation given in \citep{finn2017model} where the second and higher-order terms are neglected. We want to show a faster way to solve the \textit{\biopt{}} weight optimization problem with a tradeoff in performance. Our experimental results show that we achieve state-of-the-art performance using \sysname{}. Our results also show that \sysname{}-FO leads to a loss in performance with a commensurate gain in speed compared to the unmodified \sysname{} version. 


\begin{algorithm}[t]
\small
\caption{\sysname{}}
\begin{algorithmic}[1]
\label{alg-RWMAML}
\REQUIRE $p_{tr}, p_{val}$ distribution over training, validation tasks
\REQUIRE $m,n$ (batch sizes) and $\alpha, \eta, \gamma$ (learning rates)
\STATE Randomly initialize $\boldsymbol{\theta}$ and $W$
\WHILE{not done}
\STATE Sample mini-batch of tasks $\{\mathcal{D}_i^{S},\mathcal{D}_i^{Q}\}_{i=1}^{m} \sim p_{tr}$
\STATE Sample mini-batch of tasks $\{\mathcal{V}_j^{S},\mathcal{V}_j^{Q}\}_{j=1}^{n} \sim p_{val}$
\FOR{each task $\mathcal{T}_i,\forall i\in [1,m]$ }
\STATE Compute adapted parameters $\mathcal{A}lg(\boldsymbol{\theta},\mathcal{D}_i^{S})$ with gradient descent by Eq.~(\ref{param-adaptation}) 
\STATE Compute the gradient $\nabla_{\boldsymbol{\theta}} \mathcal{L}_i(\mathcal{A}lg(\boldsymbol{\theta},\mathcal{D}_i^{S}))$ using $\mathcal{D}_i^{Q}$
\STATE Formulate the $\boldsymbol{\theta}$ as a function of weights $\boldsymbol{\theta}_{W}^{(t)}$ by Eq.~(\ref{online-equation})
\STATE Update $\mathbf{w}_i^{(t)}$ by Eq.(\ref{weight-update}) using $\{\mathcal{V}_j^{S},\mathcal{V}_j^{Q}\}_{j=1}^{n}$
\ENDFOR
\STATE Update $\boldsymbol{\theta}^{(t+1)}$ by Eq.~(\ref{model-update}) using $\{\mathcal{D}_i^{Q}\}_{i=1}^{m}$
\ENDWHILE
\end{algorithmic}
\end{algorithm}
\setlength{\textfloatsep}{5pt}

\noindent \textbf{Weights Sharing.} The number of weight hyper-parameters in the instance-level weighting scheme correlates to the number of data instances in the query sets of the meta-training tasks. We need to determine a significant amount of hyper-parameters if the number of training tasks or data instances is enormous, which in turn affects the hyper-parameter optimization algorithm, leading to instabilities during training. Accordingly, we seek to evaluate a smaller number of hyper-parameters by sharing the weights among instances. The task-weighting scheme is an occurrence of weight sharing where we share the same weight among all the instances in the query set. Apart from the task-level weighting scheme, we try to cluster tasks based on some similarity criteria to share the same weight among all the data instances in a cluster's query sets. We likewise present a sensitive analysis in the experiment section illustrating how the number of clusters in the training tasks or instances affects the \sysname{} algorithm's performance.

\subsection{Convergence of \sysname{} Algorithm}
\vspace{-1mm}
Although the MAML algorithm's convergence rate is studied~\citep{balcan2019provable,fallah2020convergence,finn2019online}, those results do not directly hold in our case since we have a \textit{\biopt{}} optimization objective instead of standard bi-level objective of the MAML. Recall that in the case of strongly convex losses, MAML admits a convergence rate of $\mathcal{O}(1/\epsilon)$~\citep{balcan2019provable, finn2019online}. In contrast, for the non-convex case, \cite{fallah2020convergence} show a weaker convergence rate of $\mathcal{O}(1/\epsilon^2)$ to a first order stationary point. In this work, we show that \sysname{} achieves a convergence rate of $\mathcal{O}(1/\epsilon^2)$ in the case of convex losses, as long as the inner learning rate is not too high. Furthermore, we show that \sysname{} converges to a critical point of meta-validation loss and not the meta-training loss since we are optimizing the meta-validation loss in the \biopt{} setting. Table \ref{tab:convergence} shows the convergence rates of MAML and \sysname{} algorithms for strongly convex and non-convex loss functions. 

\begin{table}[htbp]
\small
    \centering
    \begin{tabular}{c|c|c}
         \toprule
         Algorithm &Strongly Convex Loss &Non-Convex Loss\\
         \hline
         MAML &$\mathcal{O}(1/\epsilon)$ &$\mathcal{O}(1/\epsilon^2)$\\ \hline
         \sysname{} & $\mathcal{O}(1/\epsilon^2)$ &Open \\ \bottomrule
    \end{tabular}
    \vspace{-3mm}
    \caption{Convergence Rates of MAML and \sysname{}}
    \label{tab:convergence}
\vspace{-5mm}
\end{table}

\begin{theorem}
\label{meta-validation-convergence}
Suppose the loss function $\small \mathcal{L}(\cdot)$ is Lipschitz smooth with constant $L$, $\mu$-strongly convex, and is a twice differential function with a $\small \rho$-bounded gradient and $\mathcal{B}$-Lipschitz Hessian. Denote $\sigma$ as the variance of drawing uniformly mini-batch sample at random. Assume that the learning rate $\small \eta_t$ satisfies $\small \eta_t = \min{(1, k/T)}$ for some $ \small k>0$ such that $k/T < 1$ and $\small \gamma_t$, $1 \leq t \leq T$, is a monotone descent sequence. Let $\small \gamma_t = \min{(\frac{1}{L}, \frac{C}{\sigma \sqrt{T}})}$ for some $\small C > 0$ such that $\small \frac{\sigma\sqrt{T}}{C} \geq L$ and $ \small \sum_{t = 0}^{\infty}\gamma_t \leq \infty$, $\sum_{t = 0}^{\infty}\gamma_t^2 \leq \infty$. Then, \sysname{} satisfies: $\small \mathbb{E}\Bigg[\left\Vert\frac{1}{N}\sum_{j=1}^{N}\nabla_W\mathcal{L}(\mathcal{A}{lg}(\boldsymbol{\theta}_W^{(t)}, \mathcal{V}_j^S), \mathcal{V}_j^Q)\right\Vert^2\Bigg] \leq \epsilon$ in $\mathcal{O}(\frac{1}{\epsilon^2})$ steps. More specifically,
\begin{equation*}
\min_{0 \leq t \le T}\mathbb{E}\Bigg[\left\Vert\frac{1}{N}\sum\nolimits_{j=1}^{N}\nabla_W\mathcal{L}(\textit{Alg}(\boldsymbol{\theta}_W^{(t)}, \mathcal{V}_j^S), \mathcal{V}_j^Q)\right\Vert^2\Bigg] \leq \mathcal{O}(\frac{1}{\sqrt{T}})
\end{equation*}
\end{theorem}
Proof is given in Appendix \ref{app:conRate}. 
The difference in convergence rates between MAML and \sysname{} is due to the additional complexity involved in solving a \textit{\biopt{}} optimization problem. The convergence analysis of \sysname{} for non-convex functions is challenging and currently unknown. Even though most deep learning problems have a non-convex landscape, the algorithms initially developed for convex cases have shown promising empirical results in non-convex cases. Under this assumption, we provide an implementation that can be generalized to any deep learning architecture in Algorithm~\ref{alg-RWMAML}. 

\begin{table*}[!htbp]
\small

    \vspace{-3mm}
    \centering
    \resizebox{15cm}{2.2cm}{
    \begin{tabular}{l|ccc|cccc}
        \toprule
        & \multicolumn{6}{c}{\textbf{5-way 3-shot}}\\
        $\mathcal{D}_{out}$ & \multicolumn{3}{c}{\textbf{SVHN}} & \multicolumn{3}{c}{\textbf{FashionMNIST}}\\
        \midrule
        OOD Ratio & 30\% & 60\% & 90\% & 30\% & 60\% & 90\%\\
        \hline 
        MAML-OOD-RM(Skyline) & 57.73$\scriptstyle{\pm 0.76}$ & 55.29$\scriptstyle{\pm 0.78}$ & 54.38$\scriptstyle{\pm 0.12}$ & 56.78$\scriptstyle{\pm 0.75}$ & 55.29$\scriptstyle{\pm 0.78}$ & 53.43$\scriptstyle{\pm 0.51}$ \\
        \hline 
        MAML & 55.41$\scriptstyle{\pm 0.75}$ & 53.93$\scriptstyle{\pm 0.76}$ & 44.10$\scriptstyle{\pm 0.68}$ & 54.65$\scriptstyle{\pm 0.77}$ & 54.52$\scriptstyle{\pm 0.76}$ & 41.52$\scriptstyle{\pm 0.74}$\\
        MMAML & 51.04$\scriptstyle{\pm 0.87}$ & 50.28$\scriptstyle{\pm 0.97}$ & 41.56$\scriptstyle{\pm 0.96}$ & 50.32$\scriptstyle{\pm 0.93}$ & 47.54$\scriptstyle{\pm 1.05}$ & 42.09$\scriptstyle{\pm 0.97}$\\
        B-TAML & 53.87$\scriptstyle{\pm 0.18}$ & 49.84$\scriptstyle{\pm 0.23}$ & 42.00$\scriptstyle{\pm 0.21}$ &  51.14$\scriptstyle{\pm 0.23}$ & 46.59$\scriptstyle{\pm 0.20}$ & 36.69$\scriptstyle{\pm 0.21}$\\
        L2R  & 47.13$\scriptstyle{\pm 0.13}$ & 40.69$\scriptstyle{\pm 0.62}$ & 47.26$\scriptstyle{\pm 0.72}$ & 33.14$\scriptstyle{\pm 0.60}$ & 44.03$\scriptstyle{\pm 0.70}$ & 33.06$\scriptstyle{\pm 0.60}$ \\
        Transductive Fine-tuning  & 55.36$\scriptstyle{\pm 0.73}$ & 54.08$\scriptstyle{\pm 0.47}$ & 45.21$\scriptstyle{\pm 0.54}$ & 55.34$\scriptstyle{\pm 0.45}$ & 51.12$\scriptstyle{\pm 0.65}$ & 47.42$\scriptstyle{\pm 0.82}$ \\
        \sysname{}-FO (ours) & 54.76$\scriptstyle{\pm 1.19}$ & 45.86$\scriptstyle{\pm 1.19}$ & 
        43.55$\scriptstyle{\pm 1.20}$ &  \textbf{57.00}$\scriptstyle{\pm 1.20}$ & 
        55.18$\scriptstyle{\pm 1.16}$ & 
        48.52$\scriptstyle{\pm 1.21}$\\
        \sysname{} (ours) & \textbf{57.12}$\scriptstyle{\pm 0.81}$& \textbf{55.66}$\scriptstyle{\pm 0.78}$& 
        \textbf{52.16}$\scriptstyle{\pm 0.76}$& 
        56.66$\scriptstyle{\pm 0.78}$& 
        \textbf{56.04}$\scriptstyle{\pm 0.79}$& 
        \textbf{49.71}$\scriptstyle{\pm 0.78}$\\
        \bottomrule
    \end{tabular}}
    \resizebox{15cm}{2.2cm}{
    \begin{tabular}{l|ccc|cccc}
        \toprule
        & \multicolumn{6}{c}{\textbf{5-way 5-shot}}\\
        $\mathcal{D}_{out}$ & \multicolumn{3}{c}{\textbf{SVHN}} & \multicolumn{3}{c}{\textbf{FashionMNIST}}\\
        \midrule
        OOD Ratio & 30\% & 60\% & 90\% & 30\% & 60\% & 90\%\\
        \hline
        MAML-OOD-RM(Skyline) & 61.89$\scriptstyle{\pm 0.69}$ & 61.31$\scriptstyle{\pm 0.75}$ & 57.79$\scriptstyle{\pm 0.69}$ & 59.83$\scriptstyle{\pm 0.76}$ & 61.31$\scriptstyle{\pm 0.75}$& 59.61$\scriptstyle{\pm 0.75}$\\
        \hline
        MAML  & 58.90$\scriptstyle{\pm 0.71}$ & 58.66$\scriptstyle{\pm 0.75}$ & 49.94$\scriptstyle{\pm 0.69}$ & 59.06$\scriptstyle{\pm 0.68}$ & 59.25$\scriptstyle{\pm 0.73}$ & 49.84$\scriptstyle{\pm 0.69}$\\
        MMAML  & 52.45$\scriptstyle{\pm 1.00}$ & 52.17$\scriptstyle{\pm 1.05}$ & 46.51$\scriptstyle{\pm 1.09}$ & 51.46$\scriptstyle{\pm 0.91}$ & 54.13$\scriptstyle{\pm 0.93}$ & 50.27$\scriptstyle{\pm 1.00}$\\
        B-TAML & 58.34$\scriptstyle{\pm 0.20}$  & 56.07$\scriptstyle{\pm 0.21}$ & 49.84$\scriptstyle{\pm 0.20}$ & 55.19$\scriptstyle{\pm 0.20}$ & 52.10$\scriptstyle{\pm 0.19}$ & 40.02$\scriptstyle{\pm 0.19}$\\
        L2R & 47.11$\scriptstyle{\pm 0.51}$  & 48.01$\scriptstyle{\pm 0.70}$ &  51.53$\scriptstyle{\pm 0.71}$ & 46.03$\scriptstyle{\pm 0.30}$ & 49.15$\scriptstyle{\pm 0.68}$ & 55.03$\scriptstyle{\pm 0.46}$\\
        Transductive Fine-tuning  & 59.16$\scriptstyle{\pm 0.76}$ & 57.84$\scriptstyle{\pm 0.58}$ & $53.64\scriptstyle{\pm 0.42}$ & 56.54$\scriptstyle{\pm 0.87}$ & 56.23$\scriptstyle{\pm 0.70}$ & 54.28$\scriptstyle{\pm 0.32}$ \\
        \sysname{}-FO (ours) & 
        57.96$\scriptstyle{\pm 0.94}$& 53.66$\scriptstyle{\pm 0.95}$& 47.58$\scriptstyle{\pm 0.96}$& 
        \textbf{60.59}$\scriptstyle{\pm 0.99}$ & 
        \textbf{60.55}$\scriptstyle{\pm 0.95}$& 
        49.23$\scriptstyle{\pm 0.98}$  \\
        \sysname{} (ours) & \textbf{60.76}$\scriptstyle{\pm 0.70}$& 
        \textbf{60.53}$\scriptstyle{\pm 0.71}$& \textbf{57.88}$\scriptstyle{\pm 0.70}$& 
        60.41$\scriptstyle{\pm 0.72}$ & 
        \textbf{60.54}$\scriptstyle{\pm 0.72}$& 
        \textbf{57.95}$\scriptstyle{\pm 0.71}$  \\
        \bottomrule
    \end{tabular}}
    \vspace{-2mm}
    \caption{Few-shot classification accuracies for the OOD experiment on various evaluation setups. \textbf{\textit{mini}-Imagenet} is used as an in-distribution dataset ($\mathcal{D}_{in}$) for all experiments. 
    }
    \label{tab:ood-classification}
    \vspace{-3mm}

\end{table*}

%% file: experiments.tex
In order to corroborate \sysname{}, we aim to study two questions:
\textbf{Q1}: Can \sysname{} be successfully applied to problems where task distribution in the training domain is partially shifted from the task distribution in the testing domain?
\textbf{Q2}: Instead of learning task weights, can \sysname{} deal with problems where data instances with noisy labels are used during the meta-training stage by learning weights in an instance-level scheme?

To answer these questions, we conduct the following experiments: (1) Mix OOD tasks with the meta-training tasks to evaluate the task-level weighting scheme of \sysname{} and (2) corrupt the labels of some training samples to evaluate the instance-level weighting scheme of \sysname{}. We follow the classification experiments in \cite{finn2017model} to do few-shot learning to evaluate both the task-level and the instance-level weighting schemes. In addition, a synthetic regression experiment is conducted for the task-level weighting scheme as well. Due to the space limitation, we list synthetic regression experiments and detailed experimental settings in Appendix \ref{app:addExp}. We performed all the experiments using PyTorch, and the code is available at \url{https://bit.ly/3s3gARc}. 


\textbf{Datasets.} We use \textbf{\textit{mini}-ImageNet}~\citep{ravi2016optimization}, \textbf{SVHN}~\citep{netzer2011reading}, \textbf{FashionMNIST}~\citep{xiao2017fashion} datasets in our experiments. For the task-level weighting scheme, \textbf{\textit{mini}-ImageNet} is considered as the ID tasks source ($\mathcal{D}_{in}$). Both the \textbf{SVHN} and the \textbf{FashionMNIST} datasets are used as OOD tasks source ($\mathcal{D}_{out}$) for \textbf{\textit{mini}-ImageNet}. For instance-level weighting, \textbf{\textit{mini}-ImageNet} is considered with corrupted labels. Additional details about datasets are given in Appendix \ref{app:exp_details}.

\subsection{Task-level Weighting for OOD Tasks}
\label{sec:exp_task-level}
\vspace{-1mm}
\textbf{Settings.} We implement image classification experiments in 5-way, 3-shot (5-shot) settings. And we use a model with similar backbone architecture given in ~\citep{vinyals2016matching,finn2017model} for all baselines. We consider  a total of 20,000 training tasks containing both ID and OOD tasks where the split of ID and OOD tasks is determined by OOD ratio(0.3, 0.6, and 0.9 in this setting). At each iteration, ID tasks and OOD tasks will be sampled according to the OOD ratio. We sample the ID tasks (meta-training, meta-validation, and meta-test) from the \textbf{\textit{mini}-ImageNet} dataset and sample OOD tasks from the \textbf{SVHN} or the \textbf{FashionMNIST} dataset.  We process all images to be of size 84$\times$84$\times$3. As mentioned before, in the task-level weighting, all the data instances in a task share the same weight, reducing the weight hyper-parameters count. To further reduce them, we use the K-means clustering method to cluster the tasks and assign a single weight value to all the same cluster tasks. 


\textbf{Baselines.} In addition to \textbf{MAML}, we have \textbf{MAML-OOD-RM} which basically removes the OOD tasks during meta-training and hence is a skyline to our model. \textbf{MMAML}~\citep{vuorio2019multimodal} leverages the strengths of meta-learners by identifying the mode of the task distribution and modulating the meta-learned prior in the parameter space. \textbf{B-TAML}~\citep{lee2020l2b} uses relocated initial parameters for new arriving tasks to handle OOD tasks. We adapted \textbf{L2R}~\citep{ren2018learning} to assign weights for different tasks and optimize these weights through stochastic gradient descent. We consider \textbf{Transductive Fine-tuning}~\citep{dhillon2019baseline} as a baseline where we finetune the parameters of the model that is obtained by adding a new classifier on top of a pre-trained deep network, which is pre-trained on support and query sets of the meta-training set, using the meta-test set's support and unlabeled query set.

\begin{table*}[t!]
    \small
    
    \centering
    \begin{center}
    \vspace{-3mm}
    \begin{tabular}{l c c c | ccc}
        \toprule
         & \multicolumn{3}{c}{\textbf{5-way 3-shot}} & \multicolumn{3}{|c}{\textbf{5-way 5-shot}}\\
        
        Noise Ratio & 20\% & 30\% & 50\% & 20\% & 30\% & 50\%  \\
        \midrule
        MAML-Noise-RM & $60.2\scriptstyle{\pm 0.02}$ & $59.35\scriptstyle{\pm 0.01}$ & $58.21\scriptstyle{\pm 0.71}$ & $61.2\scriptstyle{\pm 0.21}$ & $60.3\scriptstyle{\pm 0.32}$ & $59.1\scriptstyle{\pm 0.68}$ \\
        \hline
        MAML    & $54.8\scriptstyle{\pm 0.64}$ & $53.9\scriptstyle{\pm 1.10}$ & $51.8\scriptstyle{\pm 0.12}$ & $59.2\scriptstyle{\pm 0.28}$ & $57.6\scriptstyle{\pm 0.36}$ & $53.5\scriptstyle{\pm 0.48}$   \\
        \sysname{} (ours) & $\textbf{55.24}\scriptstyle{\pm 0.72}$ & $\textbf{54.7}\scriptstyle{\pm 1.20}$ & $\textbf{53.68}\scriptstyle{\pm 0.21}$ & $\textbf{59.6}\scriptstyle{\pm 0.54}$ & $\textbf{58.16}\scriptstyle{\pm 0.87}$ & $\textbf{55.61}\scriptstyle{\pm 1.32}$ \\
        \bottomrule 
    \end{tabular}
\vspace{-3mm}
\end{center}
\vspace{-1mm}
\caption{Test accuracies on \textbf{\textit{mini}-Imagenet} with 20\%, 30\%, and 50\% flipped noisy labels during the meta-training phase.}
\label{tab:noise}
\end{table*}

\begin{figure}[!t]
\vspace{-3mm}
    \centering
    \begin{subfigure}[b]{0.23\textwidth}
        \centering
        \includegraphics[width=\linewidth, height=2.8cm]{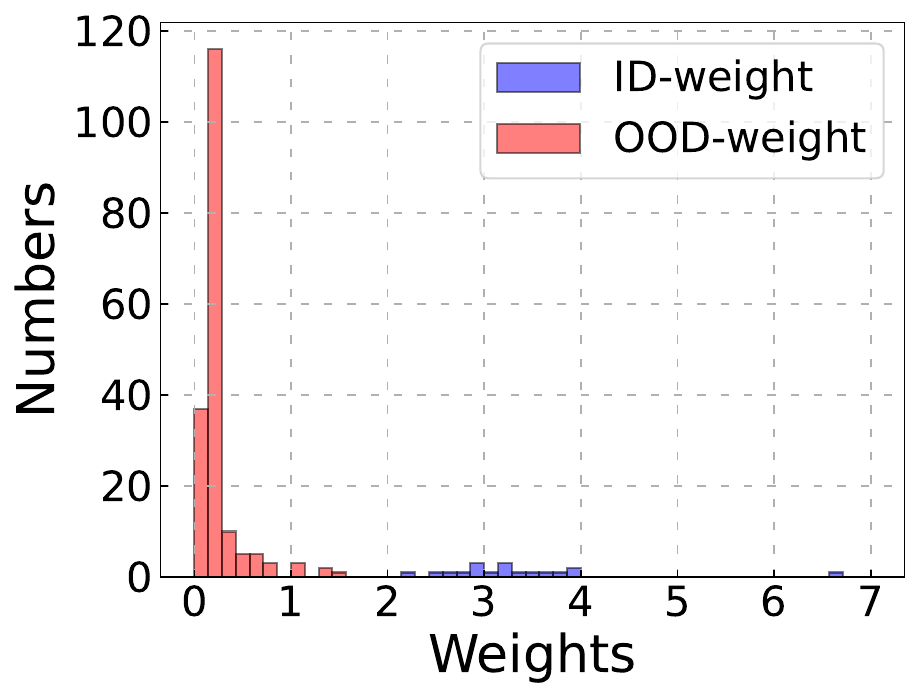}
        \caption{5-way 3-shot}
    \end{subfigure}
    \begin{subfigure}[b]{0.23\textwidth}
        \centering
        \includegraphics[width=\linewidth, height=2.8cm]{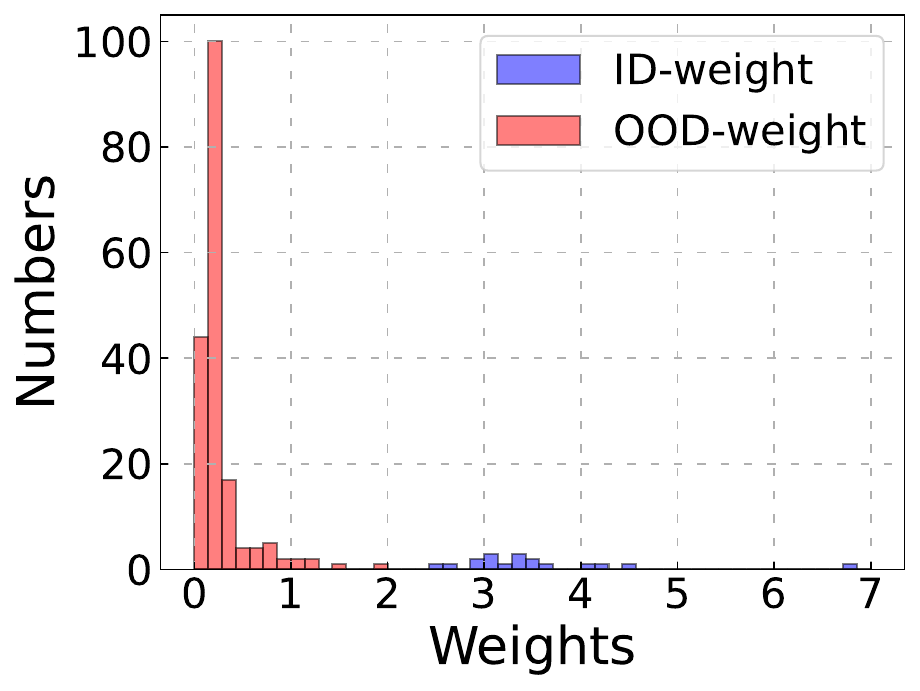}
        \caption{5-way 5-shot}
    \end{subfigure}
    \vspace{-2mm}
    \caption{Task weight distribution under 90\% ratio (SVHN).}
    \label{fig:weights histogram}
\end{figure}

\textbf{Results.} Results in Table~\ref{tab:ood-classification} show that \sysname{} significantly outperforms all baseline techniques and achieves performance competitive to the skyline method (MAML-OOD-RM) in the experiment of SVHN as OOD. For FashionMNIST OOD, \sysname{} still outperforms all baseline techniques for 60\% and 90\% ratio. For 30\% ratio, the first-order approximation, \sysname{}-FO, has the best accuracy, and \sysname{}'s accuracy is also comparable. Besides, the variance of \sysname{} is smaller than \sysname{}-FO, which means \sysname{} is more stable than \sysname{}-FO and \sysname{} still has the best performance overall. From the perspective of training time, we observed that \sysname{} takes $\mathbf{1.7 \times}$ and \sysname{}-FO takes $\mathbf{1.4 \times}$ the time taken by MAML for training. Figure~\ref{fig:weights histogram} shows weight distribution for OOD and ID tasks under 90\% ratio when SVHN is viewed as the OOD dataset for 5-way 3-shot (5-shot) settings after the meta-training phase. Both settings show that OOD tasks have much smaller weights than ID tasks: the weights belonging to OOD tasks approximately range from 0 to 1; however, the assigned weights for ID tasks are from 2 to 5, sometimes going up to 7. 

\begin{figure}[!t]
\vspace{-5mm}
    \centering
    \begin{subfigure}[b]{0.23\textwidth}
        \centering
        \includegraphics[width=\textwidth]{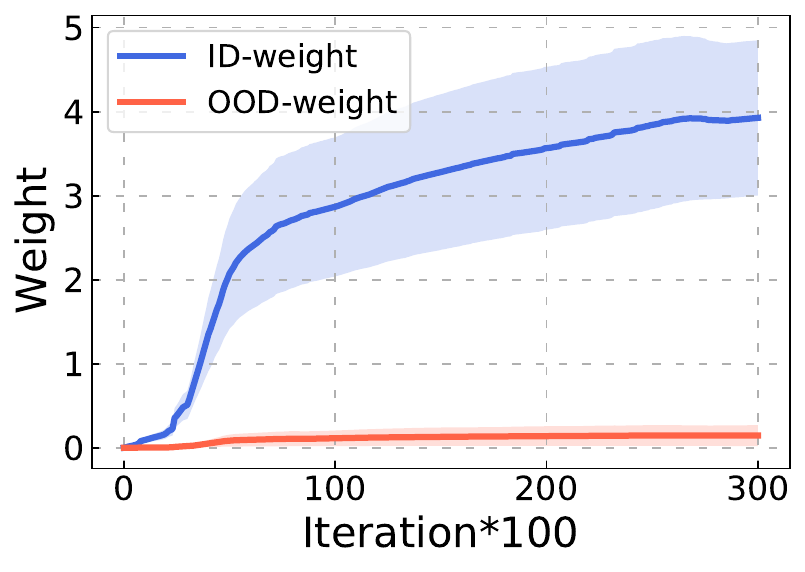}
        \caption{5-way 3-shot}
    \end{subfigure}
    \begin{subfigure}[b]{0.23\textwidth}
        \centering
        \includegraphics[width=\linewidth]{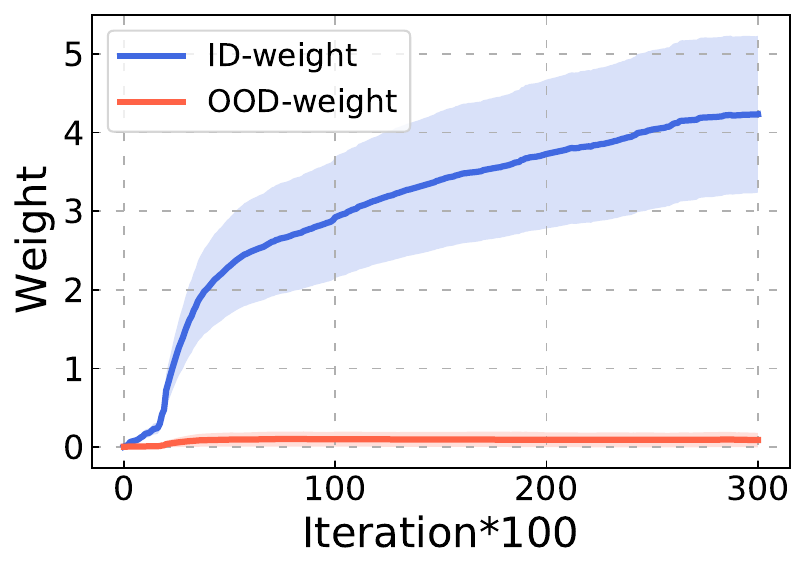}
        \caption{5-way 5-shot}
    \end{subfigure}
    \vspace{-3mm}
    \caption{Weights trend as the iterations progress for 30\% SVHN OOD experiment}
    \label{fig:svhn_wt_itera}
\end{figure}

To showcase the weights adaptation process during the training phase, we plot the weights trend as the iterations progress under the 30\% OOD ratio (SVHN) in Figure~\ref{fig:svhn_wt_itera}. 
The Blue (Red) curve denotes the mean weights for ID (OOD) tasks. The shade reflects the variance of weights. Results show that the mean weight assigned to ID tasks would increase as the iterations progress, whereas the weights assigned to OOD tasks remain close to zero, which validates the effectiveness of the \sysname{}.

\vspace{-1.5ex}

\subsection{Instance-level Weighting For Noisy Labels}
\label{sec:exp_instance-level}
\vspace{-1mm}
Similar to OOD experiments, we implement 5-way 3-shot (5-shot) experiments to evaluate the instance-level weighting scheme. We conduct experiments on noisy labels generated by randomly corrupting the original labels in \textbf{\textit{mini}-ImageNet}. Specifically, different percentages (20\%,30\%, 50\%) of training samples are selected randomly to flip their labels to simulate the noisy corrupted samples.  Intuitively, a deep model robust to noise tries to ignore the data with noisy labels. Note that data containing noisy labels only exist in the meta-training stage. Hyper-parameters are shown in Appendix~\ref{app:addExp}. 

\textbf{Baselines.} We compare our \sysname{} with the following baselines: (1) \textbf{MAML-Noise-RM} serves as a skyline. It is simply modified from MAML, and we manually fix zero weights to instances with noisy labels. (2) \textbf{MAML}.

\textbf{Results.} From the results shown in Table~\ref{tab:noise}, we can conclude that \sysname{} performs better than MAML with high accuracies. 
Furthermore, to circumvent overfitting and reduce computational complexity due to the weight matrix's high dimension, we group instance weights with 200 clusters by K-means, where instances in each cluster share the same weight initialized at 0.005. 


\begin{figure}[!t]
\vspace{-3mm}
    \centering
    \begin{subfigure}[b]{0.23\textwidth}
        \centering
         \includegraphics[height=2.9cm]{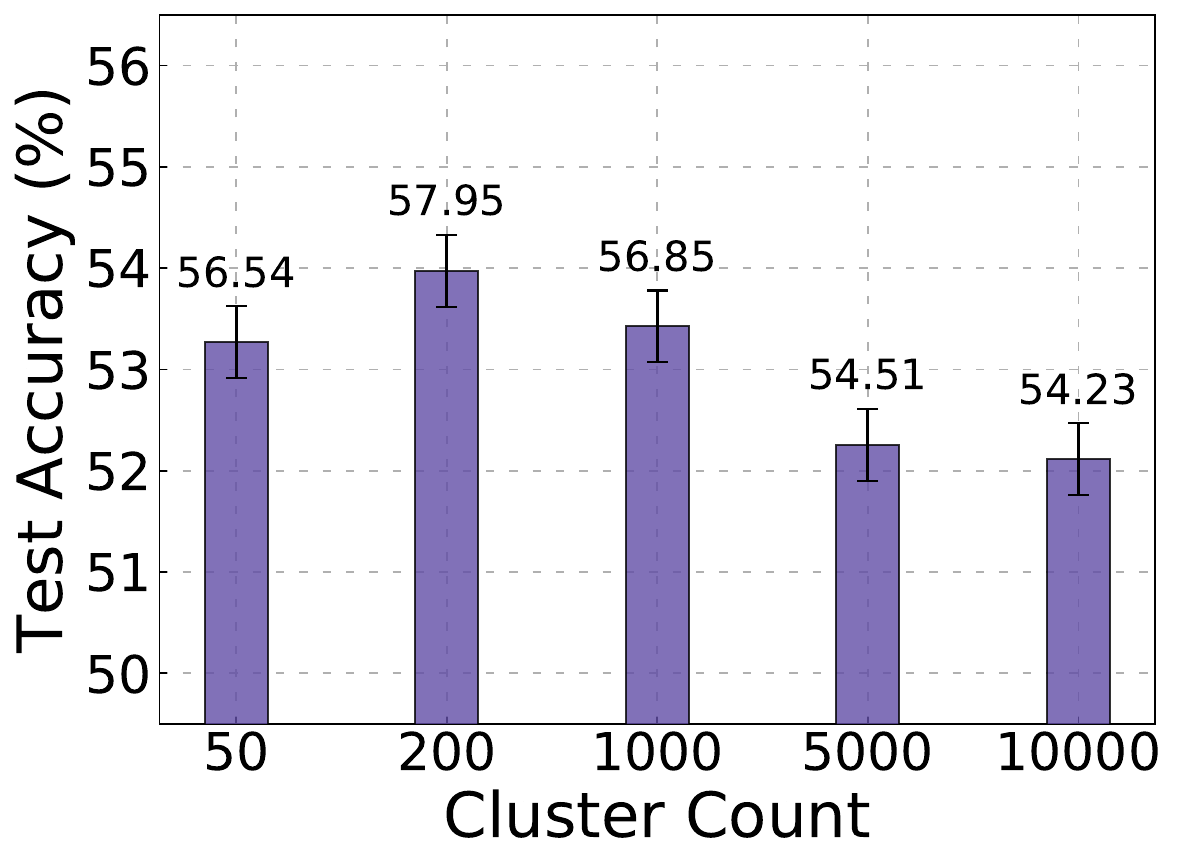}
         \caption{FashionMNIST (90\%)}
         \label{fig:cluster_ablation}
    \end{subfigure}
    \begin{subfigure}[b]{0.23\textwidth}
        \centering
         \includegraphics[height=3cm]{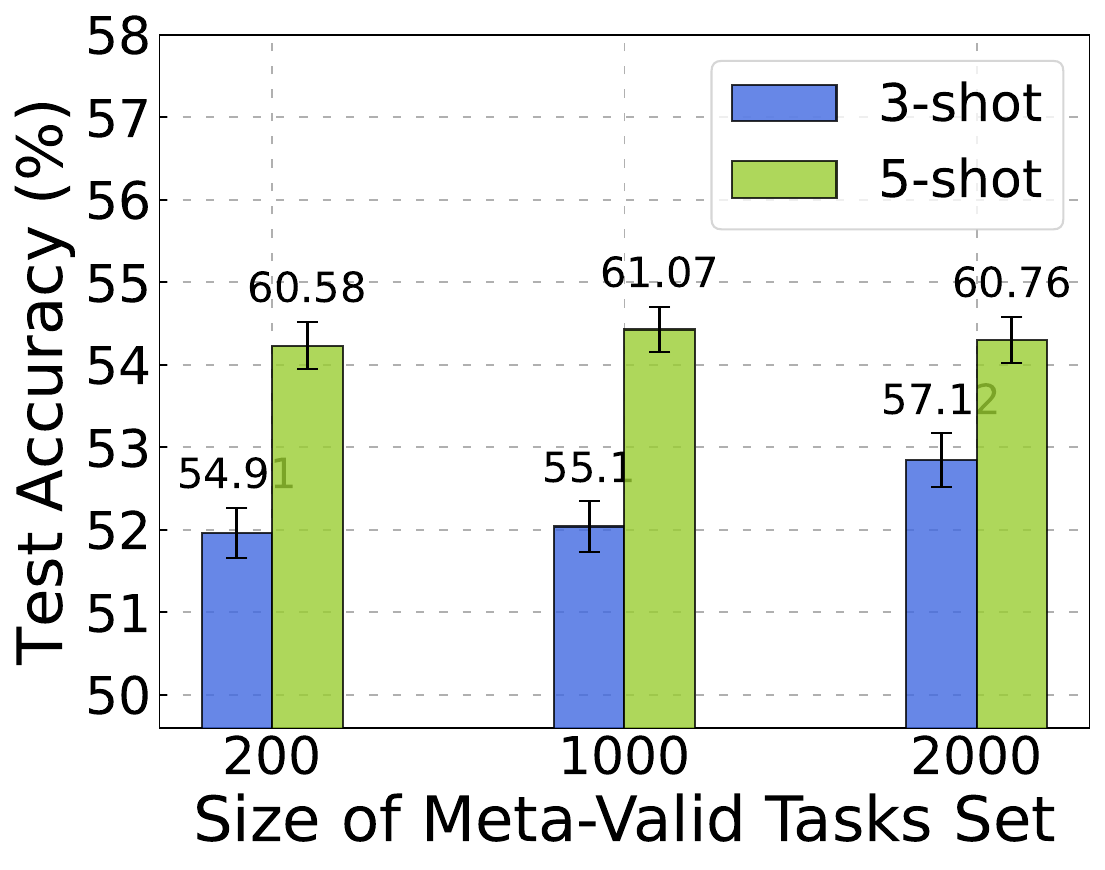}    
         \caption{SVHN (30\%)}
         \label{fig:validation_set}
    \end{subfigure}
    \vspace{-3mm}
    \caption{(a) shows accuracies under 90\% FashionMNIST OOD level with different cluster values, (b) shows accuracies under 30\% SVHN OOD level with different sizes of meta-validation tasks set.}
\end{figure}

\subsection{Sensitivity Analysis}
We perform an ablation study to determine how the number of hyper-parameters and meta-validation sets' size can affect the \sysname{} algorithm's performance. To that extent, we evaluate the \sysname{} algorithm's performance using a different number of clusters in a 5-way 5-shot 90\% FashionMNIST OOD setting. Figure~\ref{fig:cluster_ablation} shows test accuracies versus different numbers of clusters. We observed the best performance when the cluster count is 200. From Figure~(\ref{fig:cluster_ablation}), it is evident that the test accuracy decreases with an increase in the number of clusters that need to be determined. Contrarily, using a tiny number of clusters will also decrease the performance due to decreased clustering efficiency. We used 200 clusters for all our experiments. We also evaluate \sysname{} algorithm's performance using different sizes of the meta-validation set in 5-way 3-shot (5-shot) 30\% SVHN OOD setting. Figure~(\ref{fig:validation_set}) shows that \sysname{} algorithm performs well even when the meta-validation set size is tiny(i.e., 1\% of meta-training set).

%% file: conclusion.tex
\vspace{-1mm}
We propose a novel robust meta-learning algorithm for reweighting tasks/instances of corrupted data in the meta-training phase. Our method is model-agnostic, can be directly applied to any deep learning architecture in an end-to-end manner. To the best of our knowledge, \sysname{} is the first algorithm to solve a \textit{\biopt{}} optimization problem in an online manner with a convergence result. Finally, 
empirical evaluation results in OOD task and noisy label scenarios show that \sysname{} outperforms state-of-the-art meta-learning methods by efficiently mitigating the effects of unwanted instances or tasks. 

%% file: appendix.tex
\onecolumn
\begin{center}
    \Huge{Supplementary Material}
\end{center}

\section{NOTATIONS}
\label{app:notations}
For clear interpretation, we list the notations used in this paper and their corresponding explanation, as shown in Table~\ref{tab:notation}.

{\renewcommand{\arraystretch}{1.2} 
\begin{table}[ht!]
    \centering
    
    \begin{center}
    \begin{tabular}{ll}
        \toprule
        \bfseries{Notation} & \bfseries{Description}  \\
        \midrule
        $p_{tr}(\mathcal{T})$ & probability distribution of meta-training tasks  \\
        $p_{val}(\mathcal{T})$ & probability distribution of meta-validation tasks  \\
        $M, N$ & the number of meta-training, meta-validation tasks, respectively \\
        $m, n$ & batch size for $M, N$, respectively \\
        $\mathcal{T}_i$ & $i$-th meta-training task \\
        $\mathcal{T}_j^{\mathcal{V}}$ & $j$-th meta-validation task \\
        $\{\mathcal{D}_i^S$, $\mathcal{D}_i^Q\}$ & support set and query set of meta-training task $\mathcal{T}_i$ \\
        $\{\mathcal{V}_j^S$, $\mathcal{V}_j^Q\}$ & support set and query set of meta-validation task $\mathcal{T}_j^{\mathcal{V}}$\\
        $\{\boldsymbol{x}_i^k, {y}_i^k\}_{k=1}^K$ & $K$ samples in the query set $\mathcal{D}_i^Q$ of meta-training task $\mathcal{T}_i$\\
        $\boldsymbol{\theta}$  & initial parameters of base learner \\
        $\boldsymbol{\phi}_i$  & task-specific parameters for task $\mathcal{T}_i$ \\
        $\boldsymbol{\theta}_W^*$  & optimal initial parameters of base learner as a function of $W$\\
        $W$ & weight matrix for all query set samples of all meta-training tasks\\
        $\mathbf{w}_i$ & weight vector for query set samples of task $\mathcal{T}_i$\\
        $w_{ik}$ & weight for query sample $k$ for task $\mathcal{T}_i$\\
        $W^*$ & optimal weights matrix\\
        $\mathcal{L}(\boldsymbol{\phi}, \mathcal{D})$ & loss function on dataset $\mathcal{D}$ characterized by model parameter $\boldsymbol{\phi}$ \\
        $\ell(\boldsymbol{\theta}, d)$ & loss function on the query data point $d$ characterized by model parameter $\boldsymbol{\theta}$\\
        $\mathcal{A}lg(\boldsymbol{\theta}, \mathcal{D})$ & one or multiple steps of gradient descent initialized at $\boldsymbol{\theta}$ on dataset $\mathcal{D}$\\
        $\alpha, \beta, \gamma$ & step sizes \\
        \bottomrule
    \end{tabular}
\end{center}
\caption{Important Notations and Descriptions} \label{tab:notation}
\end{table}
}

\begin{itemize}
    \item $W=[\mathbf{w}_1,\mathbf{w}_2,\dots,\mathbf{w}_M]^{\intercal}$ is a matrix: $M \times K$
    \item $\mathbf{w}_i = [w_{i1},w_{i2},\dots, w_{ik}, \dots, w_{iK}]$ is a vector: $1 \times K$, weights for task $\mathcal{T}_i$
\end{itemize}

In addtion to the notations above, for notation convenience, we usually use the following notation simplicity:
$$
\mathcal{L}_i(\phi) := \mathcal{L}(\phi, \mathcal{D}_i^{Q}), \widehat{\mathcal{L}}_i(\phi) := \mathcal{L}(\phi, \mathcal{D}_i^{S})
$$
$$
\mathcal{L}_{V_j}(\phi) := \mathcal{L}(\phi, \mathcal{V}_j^{Q}), \widehat{\mathcal{L}}_{V_j}(\phi) := \mathcal{L}(\phi, \mathcal{V}_j^{S})
$$

\section{WEIGHT UPDATE OF INSTANCE AND TASK WEIGHTING SCHEME}
\label{app:Lemma1proofs}
\subsection{Proof of Lemma 1}
In this section, We restate the Lemma:~\ref{weight-update-lemma} and present the detailed proof of Lemma:~\ref{weight-update-lemma} below:
\begin{lemma*} 
The weight update for an individual weight vector $\mathbf{w}_{i}$ of the task $\mathcal{T}_i$ from time step $t$ to  $t+1$ is as follows:
\begin{align}
\label{weight-update}
    \mathbf{w}_i^{(t+1)}&=\mathbf{w}_i^{(t)} + \frac{\eta \gamma}{mn} \sum_{j=1}^{n} \nabla_{\phi_j} \mathcal{L}_{V_j}  \Big(\nabla_{\boldsymbol{\theta}}\mathcal{L}_i(\mathcal{A}lg(\boldsymbol{\theta}, \mathcal{D}_i^S))^{\intercal} \nonumber
    - \alpha  \nabla^2\widehat{\mathcal{L}}_{V_j}|_{\boldsymbol{\theta}_{W}^{(t)}} \nabla_{\boldsymbol{\theta}}\mathcal{L}_i(\mathcal{A}lg(\boldsymbol{\theta}, \mathcal{D}_i^S))^{\intercal} \Big) 
\end{align}
where $\phi_j=\mathcal{A}lg(\boldsymbol{\theta}, \mathcal{V}_{j}^{S})$.
\end{lemma*}
\begin{proof}
Our goal is to find the optimal weights by using the set of meta-validation tasks. The weight optimization objective function is as follows:
\begin{equation*}
    \begin{aligned}
     {W}^{*} &= \argmin_{\mathbf{w}} \frac{1}{N} \sum_{j=1}^{N}\mathcal{L}(\mathcal{A}lg(\boldsymbol{\theta}^{*}_{W}, \mathcal{V}_{j}^{S}), \mathcal{V}_{j}^{Q}) 
     \\
    \text{where\hspace{2mm}} \boldsymbol{\theta}^{*}_{W} &= \argmin_{\boldsymbol{\theta} \in \boldsymbol{\Theta}} \frac{1}{M}\sum_{i=1}^{M}\mathbf{w}_{i}^{*}\mathcal{L}(\mathcal{A}lg(\boldsymbol{\theta}, \mathcal{D}_i^{S}), \mathcal{D}_i^{Q})
    \end{aligned}
\end{equation*}

\textit{Remark}:
$$
\mathbf{w}_{i} = [w_{i1}, \dots, w_{iK}], \hspace{1cm} \mathcal{L}_{i}(\mathcal{A}lg(\boldsymbol{\theta}, \mathcal{D}_{i}^{S}), \mathcal{D}_{i}^{Q}) = \begin{bmatrix} \ell(\mathcal{A}lg(\boldsymbol{\theta}, \mathcal{D}_{i}^{S}), \mathcal{D}_{i1}^{Q}) \\ \dots \\ \ell(\mathcal{A}lg(\boldsymbol{\theta}, \mathcal{D}_{i}^{S}), \mathcal{D}_{ik}^{Q}) \\ \dots \\ \ell(\mathcal{A}lg(\boldsymbol{\theta}, \mathcal{D}_{i}^{S}), \mathcal{D}_{iK}^{Q}) \end{bmatrix} 
$$

Let 
\begin{equation}
F(W, \boldsymbol{\theta}) =
\frac{1}{M}\sum_{i=1}^{M}\mathbf{w}_{i}\mathcal{L}(\mathcal{A}lg(\boldsymbol{\theta}, \mathcal{D}_i^{S}), \mathcal{D}_i^{Q})
= \frac{1}{M}\sum_{i=1}^{M}\mathbf{w}_{i}\mathcal{L}_i(\mathcal{A}lg(\boldsymbol{\theta}, \mathcal{D}_i^{S}))
\end{equation}

\begin{equation}
\begin{aligned}
G(W, \boldsymbol{\theta}) = \frac{1}{N} \sum_{j=1}^{N}\mathcal{L}(\mathcal{A}lg(\boldsymbol{\theta}^{*}_{W}, \mathcal{V}_{j}^{S}), \mathcal{V}_{j}^{Q})  = \frac{1}{N} \sum_{j=1}^{N}\mathcal{L}_{V_j}(\mathcal{A}lg(\boldsymbol{\theta}^{*}_{W}, \mathcal{V}_{j}^{S}))
\end{aligned}
\end{equation}

We only consider one-step gradient update:
\begin{equation}
    \begin{aligned}
    \widehat{W} &= W - \gamma \frac{\partial G(W, \boldsymbol{\theta})}{\partial W} = W - \gamma \frac{1}{N} \sum_{j=1}^{N} \nabla_{W} \mathcal{L}_{V_j}(\mathcal{A}lg(\boldsymbol{\theta}^{*}_{W}, \mathcal{V}_{j}^{S}))
    \end{aligned}
\end{equation}
\begin{equation}
\label{theta-der}
\begin{aligned}
    \widehat{\boldsymbol{\theta}}_W &= \boldsymbol{\theta} - \eta \frac{\partial F(W, \boldsymbol{\theta})}{\partial \boldsymbol{\theta}} = \boldsymbol{\theta} - \eta \frac{1}{M}\sum_{i=1}^{M}\mathbf{w}_i \nabla_{\boldsymbol{\theta}} \mathcal{L}_{i}(\mathcal{A}lg(\boldsymbol{\theta}, \mathcal{D}_i^{S}))
    \end{aligned}
\end{equation}

\begin{equation}
    \begin{aligned}
    \nabla_{\mathbf{w}_i} \mathcal{L}_{V_j}(\mathcal{A}lg(\boldsymbol{\theta}^{*}_{\mathbf{w}}, \mathcal{V}_j^{S})) &= \frac{\partial \mathcal{L}_{V_j}}{\partial \mathcal{A}lg(\boldsymbol{\theta}_{W}^{*}, \mathcal{V}_j^{S})}\frac{\partial \mathcal{A}lg(\boldsymbol{\theta}^{*}_{W}, \mathcal{V}_j^{S})}{\partial \boldsymbol{\theta}_{W}} \frac{d \boldsymbol{\theta}_{W}}{d \mathbf{w}_i} \hspace{4cm} \\
    &= \nabla_{\phi_j} \mathcal{L}_{V_j}(\phi_j)  \frac{\partial (\boldsymbol{\theta}^{*}_{W} - \alpha \nabla \widehat{\mathcal{L}}_{V_j}(\boldsymbol{\theta}_{W}))}{\partial \boldsymbol{\theta}_{W}} (-\eta \frac{1}{M}) \nabla_{\boldsymbol{\theta}}\mathcal{L}_i(\mathcal{A}lg(\boldsymbol{\theta}, \mathcal{D}_i^{S}))^{\intercal} \\
    &= \nabla_{\phi_j} \mathcal{L}_{V_j}(\phi_j) \cdot \left(I - \alpha \nabla^2\widehat{\mathcal{L}}_{V_j}\right) (-\eta \frac{1}{M}) \nabla_{\boldsymbol{\theta}}\mathcal{L}_i(\mathcal{A}lg(\boldsymbol{\theta}, \mathcal{D}_i^{S}))^{\intercal} \\
    &= (- \frac{\eta}{M})\nabla_{\phi_j} \mathcal{L}_{V_j}(\phi_j) \cdot \left(I - \alpha \nabla^2\widehat{\mathcal{L}}_{V_j}\right)  \nabla_{\boldsymbol{\theta}}\mathcal{L}_i(\mathcal{A}lg(\boldsymbol{\theta}, \mathcal{D}_i^{S}))^{\intercal} \\
    &= -\frac{\eta}{M} \nabla_{\phi_j} \mathcal{L}_{V_j}(\phi_j) \cdot \left( \nabla_{\boldsymbol{\theta}}\mathcal{L}_i(\mathcal{A}lg(\boldsymbol{\theta}, \mathcal{D}_i^{S}))^{\intercal} - \alpha  \nabla^2\widehat{\mathcal{L}}_{V_j}\cdot \nabla_{\boldsymbol{\theta}}\mathcal{L}_i(\mathcal{A}lg(\boldsymbol{\theta}, \mathcal{D}_i^{S}))^{\intercal} \right) \\
    \end{aligned}
\end{equation}

Thus the weight update for task $\mathcal{T}_i$ can be:
\begin{align}
    \mathbf{w}_i^{(t+1)} &= \mathbf{w}_i^{(t)} - \gamma \frac{1}{n} \sum_{j=1}^{n} \nabla_{\mathbf{w}_i} \mathcal{L}_{V_j}(\mathcal{A}lg(\boldsymbol{\theta}^{*}_{W}, \mathcal{V}_j^{S}))\hspace{5cm} \nonumber\\
    &=\mathbf{w}_i^{(t)} + \frac{\eta \gamma}{mn} \sum_{j=1}^{n} \nabla_{\phi_j} \mathcal{L}_{V_j}(\phi_j)  \Big(\nabla_{\boldsymbol{\theta}}\mathcal{L}_i(\mathcal{A}lg(\boldsymbol{\theta}, \mathcal{D}_i^S))^{\intercal}
    - \alpha  \nabla^2\widehat{\mathcal{L}}_{V_j}|_{\boldsymbol{\theta}_{W}^{(t)}} \nabla_{\boldsymbol{\theta}}\mathcal{L}_i(\mathcal{A}lg(\boldsymbol{\theta}, \mathcal{D}_i^S))^{\intercal} \Big) \label{exact-noise}\\
    &\approx \mathbf{w}_i^{(t)} + \frac{\eta \gamma}{mn} \sum_{j=1}^{n} \nabla_{\phi_j} \mathcal{L}_{V_j}(\phi_j) \cdot \nabla_{\boldsymbol{\theta}}\mathcal{L}_i(\mathcal{A}lg(\boldsymbol{\theta}, \mathcal{D}_i^{S}))^{\intercal} \label{appro-noise}
\end{align}

Eq.~\eqref{exact-noise} is the exact update, namely, lemma 1. Eq.~\eqref{appro-noise} is the approximation update.
\end{proof}

\subsection{Weight Update in Task Weighting Scheme}

As mentioned in the paper, task weighting scheme is a special case of instance weighting scheme. In instance weighting scheme, each task $\mathcal{T}_i$ has a vector weight $\mathbf{w}_i$. In task weighting scheme, all query samples have the same weight. In other words, each task $\mathcal{T}_i$ has a scalar weight $w_i$. And the loss of training tasks used for the update of $\boldsymbol{\theta}$ would be the average loss for all query samples in task $\mathcal{T}_i$:
$$
\mathcal{L}_i(\mathcal{A}lg(\boldsymbol{\theta},\mathcal{D}_i^{S})) = \frac{1}{K} \sum_{k=1}^{K} \ell(\mathcal{A}lg((\boldsymbol{\theta}, \mathcal{D}_{i}^{S}), \mathcal{D}_{ik}^{Q})
$$
The weight update follows the same strategy in Lemma 1.

\section{DETAILED CONVERGENCE ANALYSIS}
\label{app:conRate}
In this section, we present the detailed proof of convergence. Before that, we first give two assumptions and several lemmas which could help for the proof of convergence.

The meta validation loss is as follows: 

$$\mathcal{L}_{V}^{meta}(\boldsymbol{\theta}^{(t)}_{W}) = \frac{1}{n}\sum_{j=1}^{n}\mathcal{L}_{V_j}(\mathcal{A}lg(\boldsymbol{\theta}^{(t)}_{W},\mathcal{V}_j^{S}))$$

Assuming that the whole weight matrix $W$ is flattened to a column matrix, the weighted meta-training loss can be written as follows:
\begin{align*}
    &\mathcal{L}_{W}^{meta}(\boldsymbol{\theta}, W) = {W}^{\intercal} \mathcal{L}_{T}^{meta}(\boldsymbol{\theta})\\
    \text{where \hspace{0.5cm}}&{\mathcal{L}}_{T}^{meta}(\boldsymbol{\theta}) = \frac{1}{m}[\mathcal{L}_{1}(\mathcal{A}lg(\boldsymbol{\theta},\mathcal{D}_1^{S})) \dots \mathcal{L}_{m}(\mathcal{A}lg(\boldsymbol{\theta},\mathcal{D}_m^{S}))]^{\intercal}
\end{align*}

The weight update equation at time step $t$ can be written as follows:
\begin{equation}
    \begin{aligned}
        {W}^{(t+1)} = {W}^{(t)} - \gamma\nabla_{W}\mathcal{L}_{V}^{meta}(\boldsymbol{\theta}^{(t)}_{W})\\
        \text{where\hspace{1cm}}\boldsymbol{\theta}^{(t)}_{W} = \boldsymbol{\theta}^{(t)} - \eta \nabla_{\boldsymbol{\theta}}\mathcal{L}_{W}^{meta}(\boldsymbol{\theta}^{(t)}, W^{(t)})\\
    \end{aligned}
    \label{online-equation-weight-1}
\end{equation}

\begin{assumption} {\upshape{($C^2$-smoothness)}} Suppose that $\mathcal{L}(\cdot)$:
\begin{itemize}
    \item is twice differentiable
    \item is $\rho$-Lipschitz in function value, i.e., $\left\Vert\nabla \mathcal{L}(\boldsymbol{\theta})\right\Vert \leq \rho$
    \item is $L$-smooth, or has $L$-Lipschitz gradients, i.e., $\left\Vert\nabla \mathcal{L}(\boldsymbol{\theta}) - \nabla \mathcal{L}(\boldsymbol{\phi})\right\Vert \leq L \left\Vert\boldsymbol{\theta} - \boldsymbol{\phi}\right\Vert$ $\forall{\boldsymbol{\theta}, \boldsymbol{\phi}}$
    \item has $\mathcal{B}$-Lipschitz hessian, i.e., $\left\Vert\nabla^2 \mathcal{L}(\boldsymbol{\theta}) - \nabla^2  \mathcal{L}(\boldsymbol{\phi})\right\Vert \leq  \mathcal{B} \left\Vert\boldsymbol{\theta} - \boldsymbol{\phi}\right\Vert$ $\forall {\boldsymbol{\theta}, \boldsymbol{\phi}}$
\end{itemize}
\end{assumption}

\begin{assumption} {\upshape{(Strong convexity)}}
  Suppose that $\mathcal{L}(\cdot)$ is convex. Further, $\mu$-strongly convex. i.e., $\left\Vert\nabla \mathcal{L}(\boldsymbol{\theta}) - \nabla \mathcal{L}(\boldsymbol{\phi})\right\Vert \geq \mu \left\Vert\boldsymbol{\theta} - \boldsymbol{\phi}\right\Vert$ $\forall{\boldsymbol{\theta}, \boldsymbol{\phi}}$\\
\end{assumption}

\begin{lemma} {\upshape{\citep{finn2019online}}} \label{online-meta-lipschitz}
     Suppose $\mathcal{L}$ and $\hat{\mathcal{L}}:$ $\mathbb{R}^d \longrightarrow \mathbb{R}$ satisfy assumptions 1 and 2. Let $\tilde{\mathcal{L}}$ be the function evaluated after a one step gradient update procedure, i.e.
    $$
\tilde{\mathcal{L}}(\boldsymbol{\theta}):= \mathcal{L}(\boldsymbol{\theta} - \alpha \nabla \hat{\mathcal{L}}(\boldsymbol{\theta}))
$$
If the step size is selected as $\alpha \leq \min{\{\frac{1}{2L}, \frac{\mu}{8\rho \mathcal{B}}\}}$, then $\tilde{\mathcal{L}}$ is convex. Furthermore, it is also $\tilde{L}=9L/8$ smooth and $\tilde{\mu}=\mu/8$ strongly convex.\\
\end{lemma}

\begin{lemma} \label{meta-lipschitz}
Suppose the loss function $\mathcal{L}$ is Lipschitz smooth with constant $L$, then the meta-validation loss $\mathcal{L}_V^{meta}$ is Lipschitz smooth with constant $\frac{9L}{8}$.\\
\end{lemma}
\begin{proof}
Since we know that,
\begin{eqnarray}
    \mathcal{L}_{V}^{meta}(\boldsymbol{\theta}) &=& \frac{1}{n}\sum_{j=1}^{n}\mathcal{L}_{V_j}(\mathcal{A}lg(\boldsymbol{\theta},\mathcal{V}_j^{S})) \nonumber\\
    &=&\frac{1}{n}\sum_{j=1}^{n}\mathcal{L}(\mathcal{A}lg(\boldsymbol{\theta},\mathcal{V}_j^{S}), \mathcal{V}_j^{Q})
\end{eqnarray}

From Lemma:~\ref{online-meta-lipschitz}, we can say that $\forall{j \in [1, n]}$, $\mathcal{L}(\mathcal{A}lg(\boldsymbol{\theta},\mathcal{V}_j^{S}), \mathcal{V}_j^{Q})$ is also lipschitz smooth with a constant of $\frac{9L}{8}$.

\begin{eqnarray}
    \left\Vert\nabla\mathcal{L}_{V}^{meta}(\boldsymbol{\theta}) - \nabla\mathcal{L}_{V}^{meta}(\boldsymbol{\phi})\right\Vert 
    &=&\left\Vert\frac{1}{n}\sum_{j=1}^{n}\bigg(\nabla\mathcal{L}(\mathcal{A}lg(\boldsymbol{\theta},\mathcal{V}_j^{S}), \mathcal{V}_j^{Q}) -\nabla \mathcal{L}(\mathcal{A}lg(\boldsymbol{\phi},\mathcal{V}_j^{S}), \mathcal{V}_j^{Q})\bigg)\right\Vert \nonumber \\
    &\leq&\frac{1}{n}\sum_{j=1}^{n}\left\Vert\bigg(\nabla\mathcal{L}(\mathcal{A}lg(\boldsymbol{\theta},\mathcal{V}_j^{S}), \mathcal{V}_j^{Q}) -\nabla \mathcal{L}(\mathcal{A}lg(\boldsymbol{\phi},\mathcal{V}_j^{S}), \mathcal{V}_j^{Q})\bigg)\right\Vert \nonumber \\
    &\leq&\frac{1}{n}\sum_{j=1}^{n}\frac{9L}{8}\left\Vert \boldsymbol{\theta} - \boldsymbol{\phi}\right\Vert \nonumber \\
    &=&\frac{9L}{8}\left\Vert \boldsymbol{\theta} - \boldsymbol{\phi}\right\Vert
\end{eqnarray}

Therefore the meta-validation loss function $\mathcal{L}_V^{meta}$ is also lipschitz smooth with constant $\frac{9L}{8}$.
\end{proof}

\begin{lemma} \label{queryloss-gradient_bound}
Suppose the loss function $\mathcal{L}$ satisfies assumption 1 and 2, then the query set loss $\mathcal{L}_{V_j}(\mathcal{A}lg(\boldsymbol{\theta},\mathcal{V}_j^{S}))$ and $\mathcal{L}_{i}(\mathcal{A}lg(\boldsymbol{\theta},\mathcal{D}_i^{S}))$  are $\rho(1+\alpha L)$-gradient bounded functions.\\
\end{lemma}
\begin{proof}
Since we know that,
\begin{eqnarray}
    \mathcal{L}_{i}(\mathcal{A}lg(\boldsymbol{\theta},\mathcal{D}_i^{S})) &=& \mathcal{L}_{i}(\boldsymbol{\theta} - \alpha\mathcal{L}(\boldsymbol{\theta}, \mathcal{D}_i^{S}))) \nonumber\\
    &=& \mathcal{L}(\boldsymbol{\theta} - \alpha\mathcal{L}(\boldsymbol{\theta}, \mathcal{D}_i^{S}), \mathcal{D}_i^{Q})
\end{eqnarray}

Suppose:
$$
\boldsymbol{\phi}=\boldsymbol{\theta}-\alpha \nabla \mathcal{L}(\boldsymbol{\theta}, \mathcal{D}_i^{S})
$$

\begin{eqnarray}
    \left\Vert\nabla\mathcal{L}_{i}(\mathcal{A}lg(\boldsymbol{\theta},\mathcal{D}_i^{S}))\right\Vert 
    &=&\left\Vert \frac{\partial \boldsymbol{\phi}}{\partial \boldsymbol{\theta}} \frac{\partial \mathcal{L}(\boldsymbol{\phi},\mathcal{D}_i^{Q})}{\partial \boldsymbol{\phi}} \right\Vert \nonumber\\
    &=&\left\Vert (1-\alpha \nabla^2\mathcal{L}(\boldsymbol{\theta},\mathcal{D}_i^S))\nabla\mathcal{L}(\phi, \mathcal{D}_i^Q) \right\Vert \nonumber\\
    &\leq&\left\Vert (1-\alpha \nabla^2\mathcal{L}(\boldsymbol{\theta},\mathcal{D}_i^S))\right\Vert \left\Vert\nabla\mathcal{L}(\phi, \mathcal{D}_i^Q) \right\Vert \nonumber\\
    &\leq& (1+\left\Vert (\alpha \nabla^2\mathcal{L}(\boldsymbol{\theta},\mathcal{D}_i^S))\right\Vert) \left\Vert\nabla\mathcal{L}(\phi, \mathcal{D}_i^Q) \right\Vert \nonumber\\
    &\leq&\rho (1+\alpha L)
\end{eqnarray}
Similarly for $\mathcal{L}_{V_j}(\mathcal{A}lg(\boldsymbol{\theta},\mathcal{V}_j^{S}))$.
\end{proof}

\begin{lemma} \label{meta-gradient_bound}
Suppose the loss function $\mathcal{L}$ is $\rho$-gradient bounded, then the meta-validation loss $\mathcal{L}_V^{meta}$ , the meta-training loss $\mathcal{L}_T^{meta}$ and the weighted meta-training loss $\mathcal{L}_W^{meta}$  are $\rho(1+\alpha L)$-gradient bounded functions.
\end{lemma}
\begin{proof}
Since we know that,
\begin{eqnarray}
    \mathcal{L}_{V}^{meta}(\boldsymbol{\theta}) &=& \frac{1}{n}\sum_{j=1}^{n}\mathcal{L}_{V_j}(\mathcal{A}lg(\boldsymbol{\theta},\mathcal{V}_j^{S})) \nonumber\\
    &=&\frac{1}{n}\sum_{j=1}^{n}\mathcal{L}(\mathcal{A}lg(\boldsymbol{\theta},\mathcal{V}_j^{S}), \mathcal{V}_j^{Q})  \nonumber
\end{eqnarray}

\begin{eqnarray}
    \left\Vert\nabla\mathcal{L}_{V}^{meta}(\boldsymbol{\theta})\right\Vert 
    &=&\left\Vert\frac{1}{n}\sum_{j=1}^{n}\nabla\mathcal{L}(\mathcal{A}lg(\boldsymbol{\theta},\mathcal{V}_j^{S}), \mathcal{V}_j^{Q})\right\Vert \nonumber \\
    &\leq&\frac{1}{n}\sum_{j=1}^{n}\left\Vert\nabla\mathcal{L}(\mathcal{A}lg(\boldsymbol{\theta},\mathcal{V}_j^{S}), \mathcal{V}_j^{Q})\right\Vert \nonumber \\
    &\leq&\frac{1}{n}\sum_{j=1}^{n}\rho(1+\alpha L) \quad \blue{\text{(From Lemma:~\ref{queryloss-gradient_bound})}}\nonumber \\
    &=&\rho(1+\alpha L)
\end{eqnarray}

Similarly, 

\begin{eqnarray}
   {\mathcal{L}}_{T}^{meta}(\boldsymbol{\theta}) &=& \frac{1}{m}[\mathcal{L}_{1}(\mathcal{A}lg(\boldsymbol{\theta},\mathcal{D}_1^{S})), \dots, \mathcal{L}_{m}(\mathcal{A}lg(\boldsymbol{\theta},\mathcal{D}_m^{S}))]^{\intercal} \nonumber \\
   &=& \frac{1}{m}[\mathcal{L}(\mathcal{A}lg(\boldsymbol{\theta},\mathcal{D}_1^{S}), \mathcal{D}_1^{S}), \dots, \mathcal{L}_{m}(\mathcal{A}lg(\boldsymbol{\theta},\mathcal{D}_m^{S}), \mathcal{D}_m^{Q})]^{\intercal}
\end{eqnarray}

\begin{eqnarray}
    \left\Vert\nabla {\mathcal{L}}_{T}^{meta}(\boldsymbol{\theta})\right\Vert &=& \left\Vert\frac{1}{m}\nabla[\mathcal{L}(\mathcal{A}lg(\boldsymbol{\theta},\mathcal{D}_1^{S}), \mathcal{D}_1^{S}), \dots, \mathcal{L}_{m}(\mathcal{A}lg(\boldsymbol{\theta},\mathcal{D}_m^{S}), \mathcal{D}_m^{Q})]^{\intercal}\right\Vert \nonumber \\
    &\leq&\frac{1}{m}\sum_{j=1}^{m}\left\Vert\nabla\mathcal{L}(\mathcal{A}lg(\boldsymbol{\theta},\mathcal{D}_j^{S}), \mathcal{D}_j^{Q})\right\Vert \nonumber \\
    &\leq&\frac{1}{m}\sum_{j=1}^{m}\rho(1+\alpha L) \quad \blue{\text{(From Lemma:~\ref{queryloss-gradient_bound})}} \nonumber \\
    &=&\rho(1+\alpha L)
\end{eqnarray}

The weighted meta-training loss is as follows:
\begin{eqnarray}
   {\mathcal{L}}_{W}^{meta}(\boldsymbol{\theta}) &=& [\mathbf{w}_1\dots\mathbf{w}_m] \cdot \frac{1}{m}[\mathcal{L}_{1}(\mathcal{A}lg(\boldsymbol{\theta},\mathcal{D}_1^{S})) \dots \mathcal{L}_{m}(\mathcal{A}lg(\boldsymbol{\theta},\mathcal{D}_m^{S}))]^{\intercal} \nonumber \\
   &=& \frac{1}{m}\sum_{i=1}^{m}\mathbf{w}_i^T\mathcal{L}_{i}(\mathcal{A}lg(\boldsymbol{\theta},\mathcal{D}_i^{S})) 
\end{eqnarray}

Since, $\mathbf{w}_i$ weight vector is normalized at every iteration such that $\left\Vert\mathbf{w}_i\right\Vert = 1$, we have :
\begin{eqnarray}
    \left\Vert\nabla {\mathcal{L}}_{W}^{meta}(\boldsymbol{\theta})\right\Vert &=& \left\Vert\frac{1}{m}\sum_{i=1}^{m}\mathbf{w}_i^T\mathcal{L}_{i}(\mathcal{A}lg(\boldsymbol{\theta},\mathcal{D}_i^{S}))\right\Vert \nonumber \\
    &\leq&\frac{1}{m}\sum_{j=1}^{m}\left\Vert\mathbf{w}_i^T \nabla\mathcal{L}_i(\mathcal{A}lg(\boldsymbol{\theta},\mathcal{D}_j^{S}))\right\Vert \nonumber \\
    &\leq&\frac{1}{m}\sum_{j=1}^{m}\left\Vert\mathbf{w}_i \right\Vert \left\Vert\nabla\mathcal{L}_i(\mathcal{A}lg(\boldsymbol{\theta},\mathcal{D}_j^{S}))\right\Vert  \nonumber\\
    &\leq&\frac{1}{m}\sum_{j=1}^{m}\left\Vert\mathcal\nabla{L}_i(\mathcal{A}lg(\boldsymbol{\theta},\mathcal{D}_j^{S}))\right\Vert \nonumber\\
    &\leq&\frac{1}{m}\sum_{j=1}^{m}\rho(1+\alpha L) \quad \blue{\text{(From Lemma:~\ref{queryloss-gradient_bound})}} \nonumber \\
    &=&\rho(1+\alpha L)
\end{eqnarray}

Therefore the meta-validation loss function $\mathcal{L}_V^{meta}$ , meta-training loss function $\mathcal{L}_T^{meta}$ and weighted meta-training loss $\mathcal{L}_W^{meta}$are $\rho(1+\alpha L)$-gradient bounded functions.
\end{proof}

\begin{lemma} \label{meta-weight-lipschitz}
Suppose the meta validation loss function $\mathcal{L}_V^{meta}$ is Lipschitz smooth with constant $L$, and the meta training loss function $\mathcal{L}_T^{meta}$ have $\rho$-bounded gradients with respect to training/validation data. Then the gradient of meta validation loss with respect to $W$ is Lipschitz continuous.
\end{lemma}

\begin{proof}
For meta approximation method, then the gradient of meta-validation loss with respect to $W$ can be written as follows:
\begin{eqnarray}
    \nabla_W \mathcal{L}_V^{meta}(\boldsymbol{\theta}_W) &=& \frac{\partial \mathcal{L}_V^{meta}(\boldsymbol{\theta}_{W})}{\partial \boldsymbol{\theta}_{W}} \cdot \frac{\partial \boldsymbol{\theta}_{W}}{\partial W} \nonumber \\
    &=& \frac{\partial \mathcal{L}_V^{meta}(\boldsymbol{\theta}_W)}{\partial \boldsymbol{\theta}_W} \cdot  \frac{\partial\left(\boldsymbol{\theta} - \eta \frac{\partial \mathcal{L}_{W}^{meta}(\boldsymbol{\theta})}{\partial \boldsymbol{\theta}}\right)}{\partial W} \nonumber \\
    &=& - \eta \frac{\partial \mathcal{L}_V^{meta}(\boldsymbol{\theta}_W)}{\partial \boldsymbol{\theta}_W} \cdot  \frac{\partial \mathcal{L}_{T}^{meta}(\boldsymbol{\theta})}{\partial \boldsymbol{\theta}} \nonumber \\
    \label{w_grad}
\end{eqnarray}
Taking gradient with respect to $W$ on both sides of Eq~\eqref{w_grad}, we have:
\begin{eqnarray}
     \|\nabla^2_W \mathcal{L}_V^{meta}(\boldsymbol{\theta}_W) \|
     &=& \eta \Big\|\frac{\partial}{\partial W} \Big( \frac{\partial \mathcal{L}_V^{meta}(\boldsymbol{\theta}_W)}{\partial \boldsymbol{\theta}_W} \cdot \frac{\partial\mathcal{L}_T^{meta}(\boldsymbol{\theta})}{\partial \boldsymbol{\theta}} \Big)\Big\| \nonumber \\
     &=& \eta \Big\|\frac{\partial\boldsymbol{\theta}_W}{\partial W} \Big( \frac{\partial^2 \mathcal{L}_V^{meta}(\boldsymbol{\theta}_W)}{\partial \boldsymbol{\theta}_W\partial \boldsymbol{\theta}_W}\Big) \cdot \frac{\partial\mathcal{L}_T^{meta}(\boldsymbol{\theta})}{\partial \boldsymbol{\theta}}\Big\|  \nonumber \\
     &=& \eta \Big\|-\eta \frac{\partial \mathcal{L}_T^{meta}(\boldsymbol{\theta})}{\partial \boldsymbol{\theta}} \Big( \frac{\partial^2 \mathcal{L}_V^{meta}(\boldsymbol{\theta}_W)}{\partial \boldsymbol{\theta}_W\partial \boldsymbol{\theta}_W}\Big) \cdot \frac{\partial\mathcal{L}_T^{meta}(\boldsymbol{\theta})}{\partial \boldsymbol{\theta}}\Big\|  \nonumber \\
       &=& \eta^2 \Big\| \frac{\partial^2 \mathcal{L}_V^{meta}(\boldsymbol{\theta}_W)}{\partial \boldsymbol{\theta}_W \partial \boldsymbol{\theta}_W} \frac{\partial\mathcal{L}_T^{meta}(\boldsymbol{\theta})}{\partial \boldsymbol{\theta}} \cdot \frac{\partial \mathcal{L}_T^{meta}(\boldsymbol{\theta})}{\partial \boldsymbol{\theta}}\Big\|  \nonumber \\
       &\le & \frac{9L\eta^2 \rho^2 {(1+\alpha L)}^2}{8} \quad \blue{\text{(From Lemma:~\ref{meta-lipschitz} and Lemma:~\ref{meta-gradient_bound})}} \nonumber\\
\end{eqnarray}
Since $\|\frac{\partial^2 \mathcal{L}^{meta}_V(\boldsymbol{\theta}_W)}{\partial \boldsymbol{\theta}_W \partial \boldsymbol{\theta}_W}\|\le \frac{9L}{8}, \|\frac{\mathcal{L}_T^{meta}(\boldsymbol{\theta})}{\partial \boldsymbol{\theta}}\| \le \rho(1+\alpha L)$. Define $\tilde{L} = \frac{9\eta^2 \rho^2 {(1+\alpha L)}^2 L}{8} $, based on Lagrange mean value theorem, we have,
\begin{eqnarray}
    \|\nabla_W \mathcal{L}_V^{meta}(\boldsymbol{\theta}_{W_i}) - \nabla_W \mathcal{L}_V^{meta}(\boldsymbol{\theta}_{W_j}) \| \le \tilde{L} \|W_i-W_j\|, \quad \text{for all } W_i, W_j
\end{eqnarray}
where $\nabla_W \mathcal{L}_{V}^{meta}(\boldsymbol{\theta}^{(t)}_{W_i}) = \frac{1}{n}\sum_{j=1}^{n}\nabla_W \mathcal{L}_{V_j}(\mathcal{A}lg(\boldsymbol{\theta}^{(t)} - \eta \nabla_{\boldsymbol{\theta}}\mathcal{L}_{W}^{meta}(\boldsymbol{\theta}^{(t)}, W_i),\mathcal{V}_j^{S}))$
\end{proof}

We restate the Theorem:~\ref{meta-validation-convergence} and present the detailed proof of Theorem:~\ref{meta-validation-convergence} below:
\begin{theorem*}
Suppose the loss function $\mathcal{L}$ is Lipschitz smooth with constant $L$ and is a differential function with a $\rho$-bounded gradient, twice differential and $\mathcal{B}$-lipschitz hessian. Assume that the learning rate $\eta_t$ satisfies $\eta_t = \min{(1, k/T)}$ for some $k>0$, such that $k/T < 1$ and $\gamma_t$, $1 \leq t \leq T$ is a monotone descent sequence, $\gamma_t = \min{(\frac{1}{L}, \frac{C}{\sigma \sqrt{T}})}$ for some $C > 0$, such that $\frac{\sigma\sqrt{T}}{C} \geq L$ and $\sum_{t = 0}^{\infty}\gamma_t \leq \infty$, $\sum_{t = 0}^{\infty}\gamma_t^2 \leq \infty$. Then \sysname{} satisfies: $\mathbb{E}\Bigg[\left\Vert\frac{1}{N}\sum_{j=1}^{N}\nabla_W\mathcal{L}(\mathcal{A}{lg}(\boldsymbol{\theta}_W^{(t)}, \mathcal{V}_j^S), \mathcal{V}_j^Q)\right\Vert^2\Bigg] \leq \epsilon$ in $\mathcal{O}(1/ \epsilon^2)$ steps. More specifically,
\begin{equation}
\underset{0 \leq t \le T}{\min}\mathbb{E}\Bigg[\left\Vert\frac{1}{N}\sum_{j=1}^{N}\nabla_W\mathcal{L}(\textit{Alg}(\boldsymbol{\theta}_W^{(t)}, \mathcal{V}_j^S), \mathcal{V}_j^Q)\right\Vert^2\Bigg] \leq \mathcal{O}(\frac{1}{\sqrt{T}})
\end{equation}

where $C$ is some constant independent of the convergence process, $\sigma$ is the variance of drawing uniformly mini-batch sample at random.
\end{theorem*}

\begin{proof}
We rewrite the weight update equation at time step $t$ (Eq. ~\ref{online-equation-weight-1}) as follows:
    \begin{align*}
        {W}^{(t+1)} = {W}^{(t)} - \gamma\nabla_{W}\mathcal{L}_{V}^{meta}(\boldsymbol{\theta}^{(t)}_{W})\\
        \text{where\hspace{1cm}}\boldsymbol{\theta}^{(t)}_{W} = \boldsymbol{\theta}^{(t)} - \eta \nabla_{\boldsymbol{\theta}}\mathcal{L}_{W}^{meta}(\boldsymbol{\theta}^{(t)}, W^{(t)})
    \end{align*}
    

Based on the update equations we can write,
\begin{align*}
\mathcal{L}_V^{meta}(\boldsymbol{\theta}^{(t+1)}) - \mathcal{L}_V^{meta}(\boldsymbol{\theta}^{(t)}) &= \mathcal{L}_V^{meta}(\boldsymbol{\theta}^{({t})} - \eta\nabla_{\boldsymbol{\theta}}\mathcal{L}_W^{meta}(\boldsymbol{\theta}^{(t)}, W^{(t)})) - \mathcal{L}_V^{meta}(\boldsymbol{\theta}^{({t-1})} - \eta\nabla_{\boldsymbol{\theta}}\mathcal{L}_W^{meta}(\boldsymbol{\theta}^{(t-1)}, W^{(t-1)}))\\
&= \bigg(\underbrace{\mathcal{L}_V^{meta}(\boldsymbol{\theta}^{(t)} - \eta\nabla_{\boldsymbol{\theta}}\mathcal{L}_W^{meta}(\boldsymbol{\theta}^{(t)}, W^{(t)})) - \mathcal{L}_V^{meta}(\boldsymbol{\theta}^{(t-1)} - \eta\nabla_{\boldsymbol{\theta}}\mathcal{L}_W^{meta}(\boldsymbol{\theta}^{(t)}, W^{(t)}))}_{\text{(a)}}\bigg) +\\
&\bigg(\underbrace{\mathcal{L}_V^{meta}(\boldsymbol{\theta}^{(t-1)} - \eta\nabla_{\boldsymbol{\theta}}\mathcal{L}_W^{meta}(\boldsymbol{\theta}^{(t)}, W^{(t)})) - \mathcal{L}_V^{meta}(\boldsymbol{\theta}^{(t-1)} - \eta\nabla_{\boldsymbol{\theta}}\mathcal{L}_W^{meta}(\boldsymbol{\theta}^{(t-1)}, W^{(t-1)}))}_{\text{(b)}}\bigg)
\end{align*}

From lemma~\ref{meta-lipschitz}, the functions $\mathcal{L}_V^{meta}(\boldsymbol{\theta})$ and $\mathcal{L}_W^{meta}(\boldsymbol{\theta})$ are lipschitz smooth with lipschitz constant $L$ provided the loss function $\mathcal{L}(\theta)$ is lipschitz smooth with lipschitz constant $L$.

For term(a) using the lipschitz smoothness property, we have:
\begin{equation}
\begin{aligned}
\mathcal{L}_V^{meta}(\boldsymbol{\theta}^{(t)} - \eta\nabla_{\boldsymbol{\theta}}\mathcal{L}_W^{meta}(\boldsymbol{\theta}^{(t)}, W^{(t)})) - \mathcal{L}_V^{meta}(\boldsymbol{\theta}^{(t-1)} - \eta\nabla_{\boldsymbol{\theta}}\mathcal{L}_W^{meta}(\boldsymbol{\theta}^{(t)}, W^{(t)})) \\
\leq (\boldsymbol{\theta}^{(t)} - \boldsymbol{\theta}^{(t-1)})^T(\nabla_{\boldsymbol{\theta}}\mathcal{L}_V^{meta}(\boldsymbol{\theta}^{(t-1)} - \eta\nabla_{\boldsymbol{\theta}}\mathcal{L}_W^{meta}(\boldsymbol{\theta}^{(t)}, W^{(t)}))) + \frac{L}{2}{\left\Vert\boldsymbol{\theta}^{(t)} - \boldsymbol{\theta}^{(t-1)}\right\Vert}^2  
\end{aligned}    
\end{equation}
Since, $\boldsymbol{\theta}^{(t)} - \boldsymbol{\theta}^{(t-1)} = -\eta \nabla_{\boldsymbol{\theta}}\mathcal{L}_W(\boldsymbol{\theta}^{(t-1)}, W^{(t)})$, $\nabla_{\boldsymbol{\theta}}\mathcal{L}_W^{meta}(\boldsymbol{\theta}, W) \leq \rho(1+\alpha L)$ and $\nabla_{\boldsymbol{\theta}} \mathcal{L}_V^{meta}(\boldsymbol{\theta}) \leq \rho(1+\alpha L)$. We have:

\begin{align}
&\hspace{0.4cm}\mathcal{L}_V^{meta}(\boldsymbol{\theta}^{(t)} - \eta\nabla_{\boldsymbol{\theta}}\mathcal{L}_W^{meta}(\boldsymbol{\theta}^{(t)}, W^{(t)})) - \mathcal{L}_V^{meta}(\boldsymbol{\theta}^{(t-1)} - \eta\nabla_{\boldsymbol{\theta}}\mathcal{L}_W^{meta}(\boldsymbol{\theta}^{(t)}, W^{(t)})) \\
&\leq (-\eta \nabla_{\boldsymbol{\theta}}\mathcal{L}_W(\boldsymbol{\theta}^{(t-1)}, W^{t}))^T(\nabla_{\boldsymbol{\theta}}\mathcal{L}_V^{meta}(\boldsymbol{\theta}^{(t-1)} - \eta\nabla_{\boldsymbol{\theta}}\mathcal{L}_W^{meta}(\boldsymbol{\theta}^{(t)}, W^{(t)}))) + \frac{L}{2}{\left\Vert-\eta \nabla_{\boldsymbol{\theta}}\mathcal{L}_W(\boldsymbol{\theta}^{(t-1)}, W^{(t)})\right\Vert}^2 \\ 
&\leq \eta {\rho}^2{(1+\alpha L)}^2 (1 + \frac{\eta L}{2})
\end{align}

For term(b) using the lipschitz smoothness property, we have:
\begin{align}
&\hspace{0.4cm}\mathcal{L}_V^{meta}(\boldsymbol{\theta}^{(t-1)} - \eta\nabla_{\boldsymbol{\theta}}\mathcal{L}_W^{meta}(\boldsymbol{\theta}^{(t)}, W^{(t)})) - \mathcal{L}_V^{meta}(\boldsymbol{\theta}^{(t-1)} - \eta\nabla_{\boldsymbol{\theta}}\mathcal{L}_W^{meta}(\boldsymbol{\theta}^{(t-1)}, W^{(t-1)})) \\
&= \mathcal{L}_V^{meta}(\theta_W^{(t)}) - \mathcal{L}_V^{meta}(\boldsymbol{\theta}_W^{(t-1)}) \\
&\leq {(W^{(t)} - W^{(t-1)})}^{T} \nabla_W \mathcal{L}_V^{meta}(\boldsymbol{\theta}_W^{(t-1)}) + \frac{\tilde{L}}{2}{\left\Vert W^{(t)} - W^{(t-1)}\right\Vert}^2 \quad \blue{\text{(From Lemma:~\ref{meta-weight-lipschitz})}} \nonumber\\
&= {-\gamma \nabla_W \mathcal{L}_V^{meta}(\boldsymbol{\theta}_W^{(t-1)})}^{T} \nabla_W \mathcal{L}_V^{meta}(\boldsymbol{\theta}_W^{(t-1)}) + \frac{\tilde{L}}{2}{\left\Vert -\gamma \nabla_W \mathcal{L}_V^{meta}(\boldsymbol{\theta}_W^{(t-1)})\right\Vert}^2\\
&= (\frac{\tilde{L}\gamma^2}{2} - \gamma){\left\Vert \nabla_W \mathcal{L}_V^{meta}(\boldsymbol{\theta}_W^{(t-1)})\right\Vert}^2
\end{align}

Combining both the inequalities for form(a) and form(b), we have:
\begin{equation}
    \begin{aligned}
        \mathcal{L}_V^{meta}(\boldsymbol{\theta}^{t+1}) - \mathcal{L}_V^{meta}(\boldsymbol{\theta}^{t}) \leq \eta {\rho}^2{(1+\alpha L)}^2 (1 + \frac{\eta L}{2}) + (\frac{\tilde{L}\gamma^2}{2} - \gamma){\left\Vert \nabla_W \mathcal{L}_V^{meta}(\boldsymbol{\theta}_W^{(t-1)})\right\Vert}^2
    \end{aligned}
\end{equation}

Summing up the above inequality from $t=1$ to $t=T-1$ and rearranging the terms, we can obtain
\begin{eqnarray}
    \sum_{t=1}^{T-1} (\gamma -\frac{\tilde{L}\gamma^2}{2}) \| \nabla_W \mathcal{L}^{meta}_V(\boldsymbol{\theta}_W^{(t)})\|^2_2 &\le& 
    \mathcal{L}_V^{meta}(\boldsymbol{\theta}^{(1)}) - \mathcal{L}_V^{meta}(\boldsymbol{\theta}^{(T)}) + \eta {\rho}^2{(1+\alpha L)}^2(\frac{\eta L(T-1)}{2}+T-1) \nonumber \\
    &\le& \mathcal{L}_V^{meta}(\boldsymbol{\theta}^{(1)}) + \eta {\rho}^2{(1+\alpha L)}^2(\frac{\eta LT}{2}+T)
\end{eqnarray}
Furthermore, we can deduce that,
\begin{eqnarray}
    \min_t \mathbb{E}\big[ \| \nabla_W  \mathcal{L}_V^{meta}(\boldsymbol{\theta}_W^{(t)})\|^2_2 \big] &\le&  \frac{ \sum_{t=1}^{T-1} (\gamma -\frac{\tilde{L}\gamma^2}{2}) \| \nabla_W  \mathcal{L}_V^{meta}(\boldsymbol{\theta}_W^{(t)})\|^2_2 }{ \sum_{t=1}^T (\gamma -\frac{\tilde{L}}{2}\gamma^2) } \nonumber \\
    &=&  \frac{ \sum_{t=1}^{T-1} (\gamma -\frac{\tilde{L}}{2}\gamma^2) \| \nabla_W  \mathcal{L}_V^{meta}(\boldsymbol{\theta}_W^{(t)})\|^2_2 }{ \sum_{t=1}^{T-1} (\gamma -\frac{\tilde{L}}{2}\gamma^2) } \nonumber \\
     &\leq&  \frac{\Big[  2\mathcal{L}_V^{meta}(\boldsymbol{\theta}_W^{(1)}) + \eta {\rho}^2{(1+\alpha L)}^2(\eta LT + 2T) \Big]}{ \sum_{t=1}^T (2\gamma -\tilde{L}\gamma^2) } \nonumber \\
      &\leq&  \frac{2\mathcal{L}_V^{meta}(\boldsymbol{\theta}_W^{(1)}) + \eta {\rho}^2{(1+\alpha L)}^2(\eta LT + 2T)}{ \sum_{t=1}^T \gamma} \nonumber \\
      &\leq& \frac{ 2\mathcal{L}_V^{meta}(\boldsymbol{\theta}_W^{(1)})}{\gamma T} + \frac{\eta {\rho}^2{(1+\alpha L)}^2( L+2)}{\gamma} \quad \blue{(\eta = \min\{1, \frac{k}{T}\} \text{ and } \eta\leq1)} \nonumber \\
      &=& \frac{ 2\mathcal{L}_V^{meta}(\boldsymbol{\theta}_W^{(1)})}{T}\max\{L, \frac{\sqrt{T}}{C}\} + \min\{1, \frac{k}{T}\} \max\{L, \frac{\sqrt{T}}{C}\}{\rho}^2{(1+\alpha L)}^2 ( L+2)\nonumber \\
      &=& \frac{ 2\mathcal{L}_V^{meta}(\boldsymbol{\theta}_W^{(1)})}{C\sqrt{T}} + \frac{k{\rho}^2{(1+\alpha L)}^2 ( L+2)}{C\sqrt{T}} = \mathcal{O}(\frac{1}{\sqrt{T}})
\end{eqnarray}
The third inequality holds for $ \sum_{t=1}^T \gamma \le \sum_{t=1}^T (2\gamma -\tilde{L}\gamma^2)$ which made us choose a functional form of $\gamma$ to be $\gamma_t = \min{(\frac{1}{L}, \frac{C}{\sigma \sqrt{T}})}$ . 

We know that,
\begin{eqnarray}
    \mathcal{L}_{V}^{meta}(\boldsymbol{\theta}^{(t)}_{W}) &=& \frac{1}{n}\sum_{j=1}^{n}\mathcal{L}_{V_j}(\mathcal{A}lg(\boldsymbol{\theta}^{(t)}_{W},\mathcal{V}_j^{S}))\\
    &=& \frac{1}{n}\sum_{j=1}^{n}\mathcal{L}(\mathcal{A}lg(\boldsymbol{\theta}^{(t)}_{W},\mathcal{V}_j^{S}), \mathcal{V}_j^{Q})
\end{eqnarray}

Therefore, we can conclude that our algorithm achieves $\underset{0\le t \le T}{\min} \mathbb{E}[\| \frac{1}{n}\sum_{j=1}^{n}\nabla_W \mathcal{L}(\mathcal{A}lg(\boldsymbol{\theta}^{(t)}_{W},\mathcal{V}_j^{S}), \mathcal{V}_j^{Q})  \|^2_2] \le \mathcal{O}(\frac{1}{\sqrt{T}}) $ in $T$ steps.
\end{proof}

\section{ADDITIONAL EXPERIMENTS}
\label{app:addExp}

\begin{figure*}[!htbp]
    \centering
    \includegraphics[width=0.96\columnwidth, height=21cm]{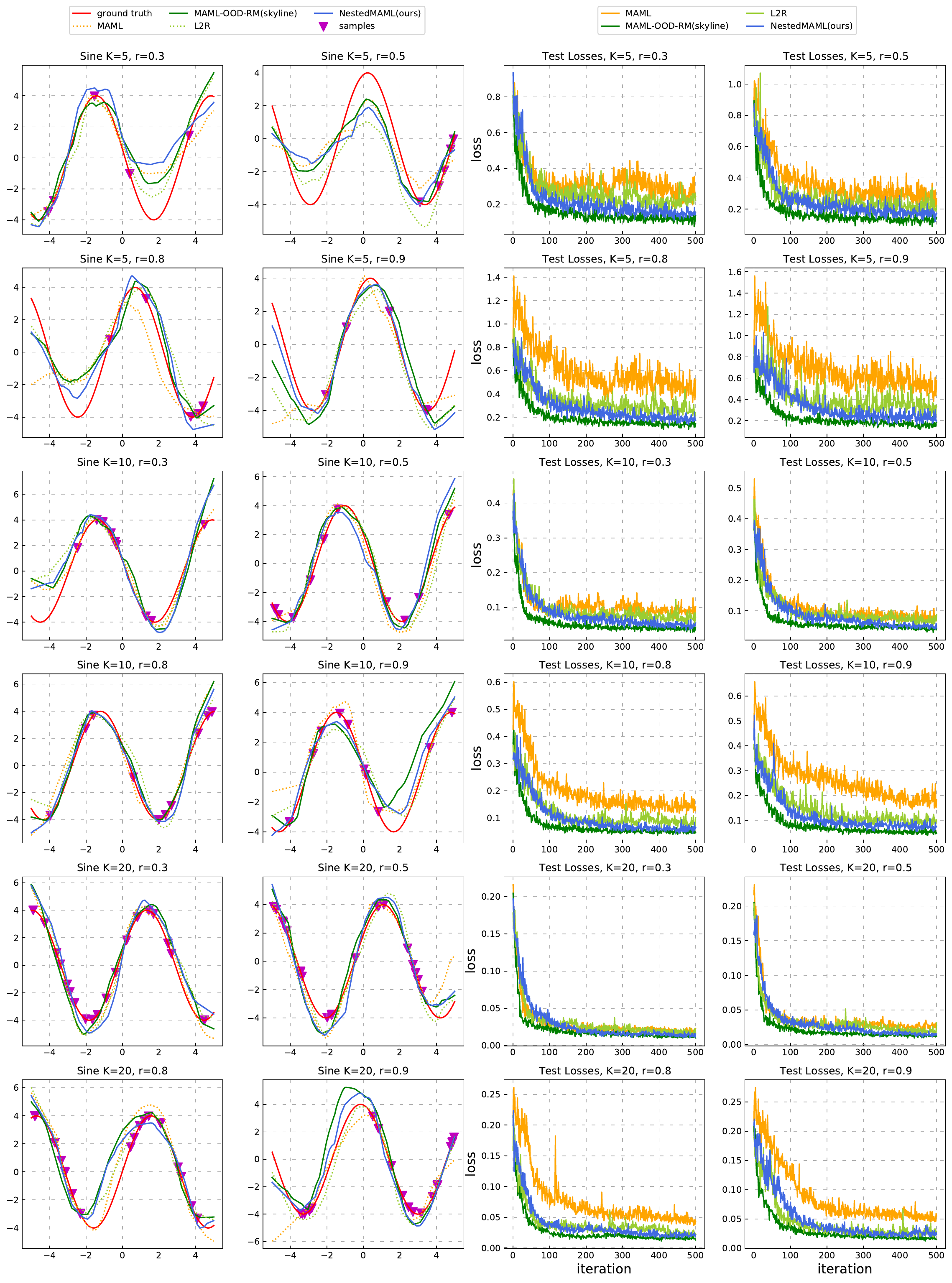}
    \caption{Results of few-shot (K=5, 10, 20) for the simple sinusoid regression task including the loss curves with respect to the number of iterations. Plotted by different levels of OOD Tasks (r = 0.3, 0.5, 0.8, 0.9).}
    \label{synthetic_regression}
\end{figure*}

\subsection{Synthetic Regression}
\textbf{Regression Setting}. To show our proposed model's robustness in the OOD task scenario, we start with a simple regression problem with outliers in the \textbf{synthetic dataset}. Specifically, during the meta-training time, each task involves $K$ samples as input and a sine wave as output, where the amplitude and phase of each sine wave are varied between tasks. More concretely, the amplitude varies within $[0.1, 5.0]$ and the phase varies within $[0, \pi]$. Datapoints from sine waves are sampled uniformly from $[-5.0, 5.0]$. In addition to in-distribution data (\textit{i.e.} data points sampled from sine waves), outliers or data points out of sine distributions (\textit{i.e.} OOD) are added into meta-training stage. To generate OOD data, we set outputs that are linear to the corresponding inputs. It is notable that, during meta-val and meta-test stages, all tasks are without any outliers. Our proposed model's intuition behind such a setting learns weights based on validation tasks and will assign higher weights to sinusoid tasks in meta-training, which could have better results. Instead, MAML uses equal weights for each meta-training task, which may not generalize good performance to unseen tasks when OOD is mixed during training. The loss function of Mean Squared Error (MSE) between prediction and the true value is applied for optimization. During meta-validation/test time, all tasks are without any outliers. Intuition: our model could learn weights based on validation tasks (sine wave) and assign higher weights to sinusoid tasks in meta-training tasks, resulting in better results. Instead, MAML uses the same weights for each task in meta-training tasks, which will not have good generalization results. 

Figure~\ref{synthetic_regression} shows the results of the \sysname{} and other baselines: \textbf{MAML}~\citep{finn2017model} and \textbf{L2R} ~\citep{ren2018learning}. The baseline \textbf{MAML-OOD-RM} corresponds to a MAML model trained just on In-Distribution (ID) tasks and will act as skyline. From Figure~\ref{synthetic_regression} and Table~\ref{tab:regres}, it is evident that \sysname{} algorithm performs better than other baseline methods and achieved low MSE error values.

\begin{table*}[!htbp]
\small
    \centering
\begin{tabular}{c|l|c|c|c|c}
    \toprule
    Shots K & Methods & r=0.3 & r=0.5 & r=0.8 & r=0.9   \\
    \midrule
    \multirow{4}{*}{5} & MAML-OOD-RM(skyline)  & 0.1357 & 0.1460 & 0.1457 & 0.1830 \\
    \cline{2-6} 
    & MAML  & 0.2448  & 0.2658 & 0.5200 & 0.5807 \\
    & L2R   & 0.2228  & 0.2225 & 0.2137 & 0.3361 \\
    & \sysname{} (ours)  & \textbf{0.1548} & \textbf{0.1725} & \textbf{0.1761} & \textbf{0.1971} \\
    \midrule
    \multirow{4}{*}{10} & MAML-OOD-RM(skyline)  & 0.0430 & 0.0425 & 0.0466 & 0.0485 \\
    \cline{2-6} 
    & MAML  & 0.1015 & 0.0865 & 0.1397 &  0.1831 \\
    & L2R   & 0.0723 & 0.0888 & 0.1022 &  0.0978 \\
    & \sysname{} (ours)  & \textbf{0.0552} & \textbf{0.0458} & \textbf{0.0653} & \textbf{0.0743} \\
    \midrule
    \multirow{4}{*}{20} & MAML-OOD-RM(skyline)  & 0.0102 & 0.0120  & 0.0131 & 0.0150 \\
    \cline{2-6} 
    & MAML  & 0.0228 &  0.0278  & 0.0432 & 0.0553 \\
    & L2R   & 0.0169 &  0.0314  & 0.0219 & 0.0289 \\
    & \sysname{} (ours)  &  \textbf{0.0152} & \textbf{0.0153} & \textbf{0.0221} & \textbf{0.0231} \\
    \bottomrule
\end{tabular}
\caption{MSE loss for the OOD experiment on various evaluation setups. \textbf{sinusoid} is used as an in-distribution dataset ($\mathcal{D}_{in}$) for all experiments. }
\label{tab:regres}

\end{table*}

\subsection{More Experimental Details} \label{app:exp_details}

\paragraph{Datasets: }\textbf{\textit{Mini}-ImageNet}~\citep{ravi2016optimization} contains 60,000 images of size $84\times84\times3$ from 100 classes. We use the split proposed in~\cite{ravi2016optimization}: 64 classes for training, 12 classes for validation and 24 classes for testing. \textbf{SVHN}~\citep{netzer2011reading}, a street view house numbers dataset, contains 26,032 images of  size $32\times32\times3$ from 10 digits classes. \textbf{FashionMNIST}~\citep{xiao2017fashion}, a fashion dataset(\textit{i.e.} clothes, shoes, \textit{etc}), contains 60,000 grayscale images of size $28\times28$ pixels from 10 classes.


\paragraph{Details of Settings: } As aforementioned, our backbone follows the same architecture as the embedding function used by \citep{finn2017model}. Specially, the backbone structure consists of 4 modules, each of which contains a $3\times3$ convolutions and 64 filters, followed by batch normalization, a ReLU, and a $2\times2$ max-pooling with stride 2. To reduce overfitting, 32 filters per layer are considered. We use the same model for OOD and ID tasks during the meta-training stage, so it's necessary to make sure the image sizes are consistent. We resize the image size of SVHN and FashionMNIST to $84\times84\times3$ which is consistent with \textit{mini}-ImageNet when evaluating the task-level weighting scheme. We also use the same backbone when evaluating the instance-level weighting scheme. Cross entropy loss function is used for these two schemes.

\paragraph{Parameter Tuning for Task-level Scheme: }All baseline approaches follow the original implementation including hyper-parameters. For our \sysname{} algorithm, all step sizes ($\alpha, \eta, \gamma$) are chosen from $\{0.0001, 0.001, 0.01, 0.1\}$. Batch size ($m,n$) are chose from $\{4, 10, 20, 25, 32\}$. The number of iterations are chosen from $\{10,000,\enspace 20,000,\enspace 30,000,\enspace 40,000,\enspace 60,000\}$. The number of clusters used in K-means is chosen from $\{50,\enspace 200,\enspace 1,000,\enspace 5,000,\enspace 10,000\}$. The selected best ones are: Fast model parameters step size $\alpha=0.01$, meta parameters step size $\eta=0.001$, weight update step size $\gamma=0.1$; mini-batch size $m=n=10$; the number of iterations in $30\%,60\%$ are $30,000$, $90\%$ is $60,000$ respectively. The number of clusters is 200. 

\paragraph{Parameter Tuning for Instance-level Scheme: }Tuning hyper-parameters follows the same aforementioned strategy. The selected best ones are: Fast model parameters step size $\alpha=0.01$, meta parameters step size $\eta=0.001$, weight update step size $\gamma=0.01$; mini-batch size $m=n=10$; the number of iterations in $20\%,30\%,50\%$ are $20,000$. The number of clusters is 200.  

Other related hyperparameters are kept the same with MAML. For example, 5 gradient steps are used when training the backbone in these two schemes, and 10 gradient steps during the meta-test stage. The number of instances in the query set of each task is 15.